\documentclass{article} %
\usepackage[authoryear,sort]{natbib}
\usepackage[bitstream-charter]{mathdesign}
\usepackage[
  paper  = letterpaper,
  left   = 1.2in,
  right  = 1.2in,
  top    = 1.0in,
  bottom = 1.0in,
  ]{geometry}

\usepackage{amsthm,amsmath}

\usepackage[english]{babel}
\usepackage[parfill]{parskip}
\usepackage{afterpage}
\usepackage{framed}

\usepackage{graphicx}
\usepackage[labelfont=bf]{caption}
\usepackage{booktabs}

\usepackage{listings}
\usepackage{fancyvrb}
\fvset{fontsize=\normalsize}

\usepackage[colorlinks,
            linkcolor = blue,
            urlcolor  = blue,
            citecolor = blue,
            linktoc=all]{hyperref}

\usepackage[nameinlink]{cleveref}

\usepackage[acronym,nowarn,nohypertypes={acronym,notation}]{glossaries}
\setacronymstyle{long-sc-short}

\lstdefinestyle{mystyle}{
    commentstyle=\color{OliveGreen},
    keywordstyle=\color{BurntOrange},
    numberstyle=\tiny\color{black!60},
    stringstyle=\color{MidnightBlue},
    basicstyle=\ttfamily,
    breakatwhitespace=false,
    breaklines=true,
    captionpos=b,
    keepspaces=true,
    numbers=left,
    numbersep=5pt,
    showspaces=false,
    showstringspaces=false,
    showtabs=false,
    tabsize=2
}
\usepackage{tikz}
\usetikzlibrary{shapes,decorations,arrows,calc,arrows.meta,fit,positioning}
\tikzset{
    -Latex,auto,node distance =1 cm and 1 cm,semithick,
    state/.style ={circle, draw, minimum width = 0.7 cm},
    detstate/.style ={rectangle, draw, minimum width = 0.7 cm, minimum height = 0.7 cm},
    point/.style = {circle, draw, inner sep=0.04cm,fill,node contents={}},
    bidirected/.style={Latex-Latex,dashed},
    el/.style = {inner sep=2pt, align=left, sloped}
}

\usepackage{wrapfig}
\usepackage{mathtools}

\usepackage{mfirstuc}
\usepackage{xspace}

\makeatletter
\def\adl@drawiv#1#2#3{%
        \hskip.5\tabcolsep
        \xleaders#3{#2.5\@tempdimb #1{1}#2.5\@tempdimb}%
                #2\z@ plus1fil minus1fil\relax
        \hskip.5\tabcolsep}
\newcommand{\cdashlinelr}[1]{%
  \noalign{\vskip\aboverulesep
           \global\let\@dashdrawstore\adl@draw
           \global\let\adl@draw\adl@drawiv}
  \cdashline{#1}
  \noalign{\global\let\adl@draw\@dashdrawstore
           \vskip\belowrulesep}}
\makeatother

\usepackage[algoruled, algo2e]{algorithm2e}
\usepackage{algorithm}
\usepackage{algorithmic}

\newacronym{smc}{\textsc{smc}}{sequential Monte Carlo}
\newacronym{dmc}{\textsc{dmc}}{diffusion Monte Carlo}
\newacronym{rlhf}{\textsc{rlhf}}{reinforcement learning from human feedback}
\newacronym{mdlm}{\textsc{mdlm}}{masked diffusion language model}
\newacronym{ips}{\textsc{ips}}{interacting particle system}
\newacronym{svdd}{\textsc{svdd}}{soft value-based decoding in diffusion models}

\newacronym{fk-ips}{\textsc{fk-ips}}{Feynman-Kac interacting particle system}
\newacronym{is}{\textsc{is}}{importance sampling}
\newacronym{tds}{\textsc{tds}}{twisted diffusion sampler}
\newacronym{method}{\textsc{fk} \MakeLowercase{steering}}{Feynman-Kac diffusion steering}

\newacronym{fk}{\textsc{fk}}{Feynman-Kac}
\newacronym{fid}{\textsc{fid}}{Frechet inception distance}
\newacronym{ssim}{\textsc{ssim}}{structural similarity metric}

\newacronym{sde}{\textsc{sde}}{stochastic differential equation}
\newacronym{elbo}{\textsc{elbo}}{evidence lower bound}
\newacronym{kl}{\textsc{kl}}{Kullback-Leibler}

\newacronym{bpd}{\textsc{bpd}}{bits-per-dim}
\newacronym{vae}{vae}{variational autoencoder}

\newacronym{ode}{\textsc{ode}}{ordinary differential equation}
\newacronym{dbgm}{\textsc{dbgm}}{diffusion-based generative model}

\newacronym{vpsde}{\textsc{vpsde}}{variance-preserving stochastic differential equation}
\newacronym{vesde}{vesde}{variance-exploding stochastic differential equation}
\newacronym{alda}{alda}{accelerated Langevin diffusion}
\newacronym{malda}{malda}{modified accelerated Langevin diffusion}

\newacronym{mle}{mle}{maximum likelihood estimation}

\newacronym{cdf}{cdf}{cumulative density function}

\newacronym{hsm}{hsm}{Hybrid Score Matching}
\newacronym{dsm}{dsm}{Denoising Score Matching}
\newacronym{ism}{ism}{Implicit Score Matching}

\newacronym{iwae}{iwae}{importance-weighted auto-encoder}

\newacronym{vp}{vp}{variance preserving}
\newacronym{ve}{ve}{variance exploding}

\newacronym{mdm}{mdm}{Multivariate Diffusion Model}

\newacronym{pfode}{pfode}{\textit{probability flow} \gls{ode}}

\newacronym{fpe}{fpe}{Fokker-Planck equation}

\newcommand{\qdata}{q_{\text{data}}}
\newcommand{\pfk}[1]{p_{\textsc{fk}, #1}}
\newcommand{\prior}{\pi_{\text{prior}}}

\newcommand{\ptarget}{p_{\text{target}}}

\newcommand{\g}{\,|\,}

\newtheorem*{assumption*}{Assumption}

\newtheorem*{corollary*}{Corollary}

\newtheorem*{definition*}{Definition}

\newtheorem*{lemma*}{Lemma}

\newtheorem*{proposition*}{Proposition}

\newtheorem*{theorem*}{Theorem}

\DeclareMathOperator*{\argmin}{arg\,min}

\renewcommand{\mid}{~\vert~}

\newcommand{\norm}[1]{\left\lVert#1\right\rVert}

\newcommand{\mbc}{\mathbf{c}}

\newcommand{\mbx}{\mathbf{x}}

\newcommand{\mbZ}{\mathbf{Z}}

\newcommand{\cN}{\mathcal{N}}

\newcommand{\E}{\mathop{\mathbb{E}}} %

\usepackage{nicefrac}
\usepackage{subcaption}
\usepackage{wrapfig}
\usepackage{titlesec}

\usepackage[utf8]{inputenc} 
\usepackage[T1]{fontenc}    
\usepackage{hyperref}       
\usepackage{url}            
\usepackage{booktabs}       
\usepackage{nicefrac}       
\usepackage{xcolor}        
\usepackage{soul}
\usepackage{float}

\usepackage{hyperref}
\usepackage{url}
\usepackage{enumitem}
\usepackage{natbib}

\title{A General Framework for Inference-time Scaling and Steering of Diffusion Models}

\author{
 Raghav Singhal$^{*,1}$, Zachary Horvitz$^{*,2}$, Ryan Teehan$^{*,3}$ \\ 
Mengye Ren$^{1,3}$, Zhou Yu$^{2}$, Kathleen McKeown$^{2}$,  Rajesh Ranganath$^{1,3}$
     \\\\
     \hspace{-10pt} 
     \textsuperscript{1}Department of Computer Science, New York University  \\
     \hspace{-10pt}           \textsuperscript{2}Columbia University  \\
     \hspace{-10pt} 
     \textsuperscript{3}Center for Data Science, New York University \\
     \hspace{-10pt} 
}
\date{}

\begin{document}

\maketitle

\begin{abstract}
    Diffusion models produce impressive results in modalities ranging from images and video to protein design and text. However, generating samples with user-specified properties remains a challenge. 
    Recent research proposes fine-tuning models to maximize rewards that capture desired properties, but these methods require expensive training and are prone to mode collapse.    
In this work, we present \gls{fk} steering, an inference-time framework for steering diffusion models with reward functions. \gls{fk} steering works by sampling a system of multiple interacting diffusion processes, called \textit{particles}, and resampling particles at intermediate steps based on scores computed using functions called \textit{potentials}. Potentials are defined using rewards for intermediate states and are selected such that a high value indicates that the particle will yield a high-reward sample.
    We explore various choices of potentials, intermediate rewards, and samplers. We evaluate \gls{fk} steering on text-to-image and text diffusion models.
    For steering text-to-image models with a human preference reward, we find that \gls{fk} steering a 0.8B parameter model outperforms a 2.6B parameter fine-tuned model  on prompt fidelity, with \textit{faster sampling and no training}. For steering text diffusion models with rewards for text quality and specific text attributes, we find that \gls{fk} steering generates lower perplexity, more linguistically acceptable outputs and enables gradient-free control of attributes like toxicity. 
    Our results demonstrate that inference-time scaling and steering of diffusion models -- even with off-the-shelf rewards -- can provide significant sample quality gains and controllability benefits. Code is available \href{https://github.com/zacharyhorvitz/FK-Diffusion-Steering}{here}. \def\thefootnote{*}\footnotetext{Denotes equal authorship. Correspondence to \url{rsinghal@nyu.edu}, \url{zfh2000@columbia.edu}, \url{rst306@nyu.edu}.}
\end{abstract}

\begin{figure}[t]
    \centering
    \includegraphics[width=0.9\linewidth]{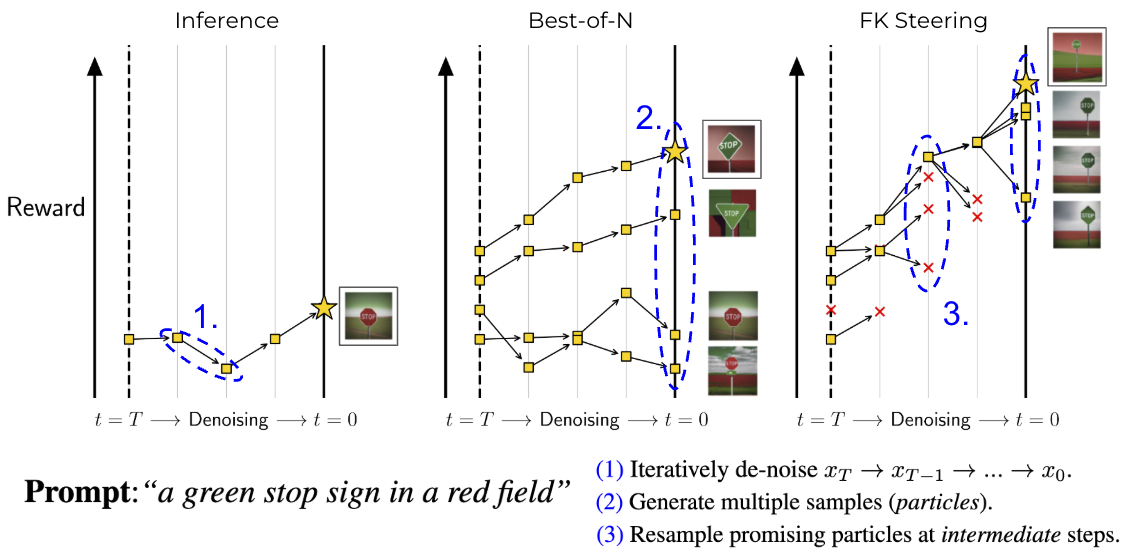}
    \caption{
    \textbf{\gls{method}} is a particle-based steering approach that generates multiple samples (\textit{particles}) like best-of-$n$ (importance sampling) approaches. In \gls{method}, particles are evaluated at \textit{intermediate} steps, where they are scored with functions called \textit{potentials}. Potentials are defined using intermediate rewards and are selected such that promising particles are resampled and poor samples are terminated. Additionally, potentials are selected such that resulting outputs are samples from the tilted distribution, $\mbx_0 \propto p_\theta(\mbx_0) \exp \left(\lambda r(\mbx_0)\right).$
    }
    \label{fig:fk_graphic}
\end{figure}

\section{Introduction}

Diffusion-based generative models \citep{sohl2015deep} have led to advances in modeling images \citep{ho2020denoising, song2020score}, videos \citep{ho2022video}, and proteins \citep{gruver2023proteindesignguideddiscrete}, as well as promising results for text generation \citep{li2022diffusionlmimprovescontrollabletext, han2023ssdlmsemiautoregressivesimplexbaseddiffusion, gong2023diffuseqsequencesequencetext, gulrajani2023likelihoodbaseddiffusionlanguagemodels,horvitz2024paraguideguideddiffusionparaphrasers}. 
Despite these advances, diffusion models have failure modes. For example, there is a high failure rate for text-to-image models in terms of adherence to text prompts \citep{ghosh2024geneval}. Additionally, adapting these models to produce samples that conform to user preferences remains a challenge.

One approach for making generative models adhere to user preferences is to encode preference via a reward function $r(\mbx_0)$ and sample from a \textit{tilted} distribution $p_\text{target}(\mbx_0) \propto p_\theta(\mbx_0) \exp(r(\mbx_0))$ \citep{korbak2022rl}, such that high-reward samples are up-weighted and low-reward samples are down-weighted. These reward functions can be human preference scores \citep{xu2024imagereward,wu2023human}, vision-language models \citep{liu2024visual}
to score prompt fidelity,  or likelihoods $p(y \mid \mbx_0)$ for an attribute $y$ \citep{wu2024practical}. Current approaches for sampling from this tilted distribution can be categorized into (a) fine-tuning and (b) inference-time steering methods. 

\citet{black2023training,domingo2024adjoint,wallace2024diffusion} explore 
fine-tuning of diffusion models with reward functions. 
However, fine-tuning requires expensive training and ties the model to a single reward. 
Alternatively, two common inference-time steering approaches are reward gradient-based guidance \citep{song2020score,bansal2023universal} and best-of-$n$ sampling. Best-of-$n$ sampling can be used to guide any diffusion model with generic reward functions, however, it allocates a large amount of computation to samples that yield low rewards \citep{chatterjee2018sample}. Gradient-based guidance presents an efficient alternative, but it is limited to differentiable reward functions and continuous-state diffusion models. Therefore, steering a diffusion model at inference-time with arbitrary rewards remains a challenge.

In this work, we present Feynman-Kac steering (\gls{method}), a flexible framework for steering diffusion-based generative models with arbitrary rewards that uses  \gls{fk} interacting particle system methods \citep{moral2004feynman,vestal2008interacting}. We generalize previous works that define Feynman-Kac measures to conditionally sample diffusion models \citep{trippe2022diffusion,wu2024practical,chung2022diffusion,janati2024divide}. \gls{method} enables guidance with arbitrary reward functions, differentiable or otherwise, for both discrete and continuous-state models. The approach makes use of a rare-event simulation method, \gls{fk-ips} \citep{moral2004feynman,del2005genealogical,hairer2014improved,vestal2008interacting}. \gls{fk-ips} enables the generation of samples with high-rewards, which may be rare events under the original model $p_\theta(\mbx)$.

\begin{figure*}[t]
    \centering
    \includegraphics[width=0.6\linewidth]{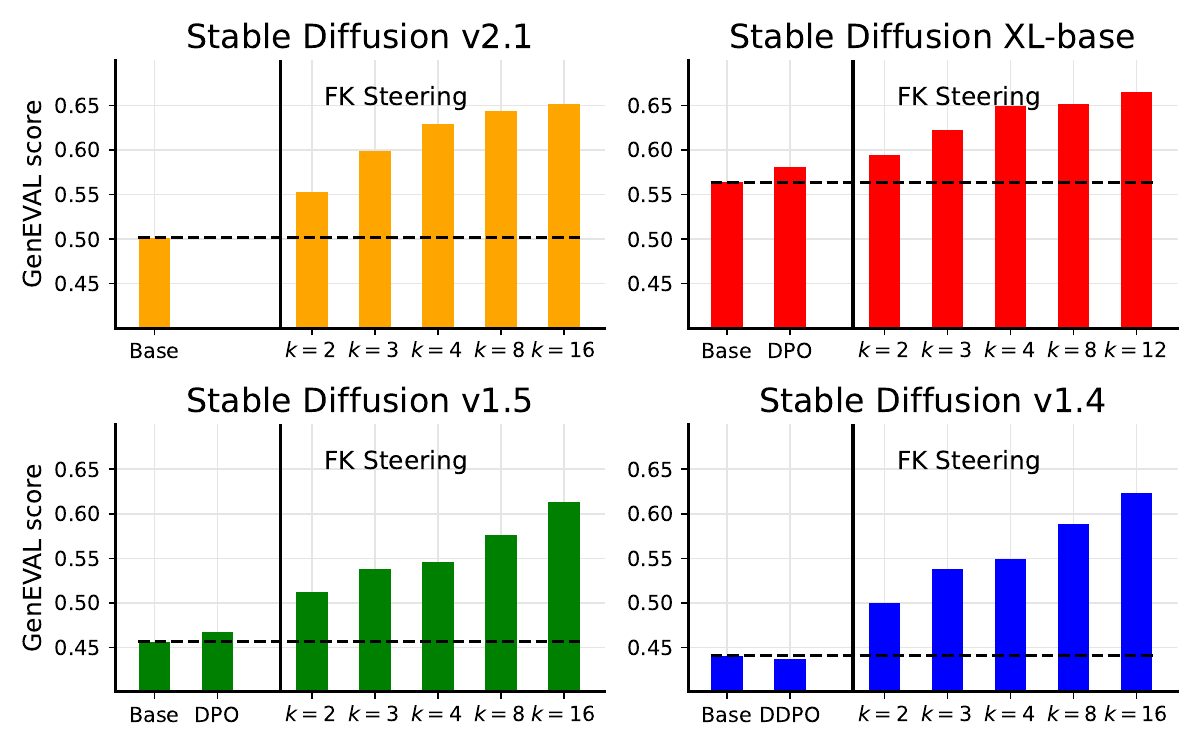}\includegraphics[width=0.4\linewidth]{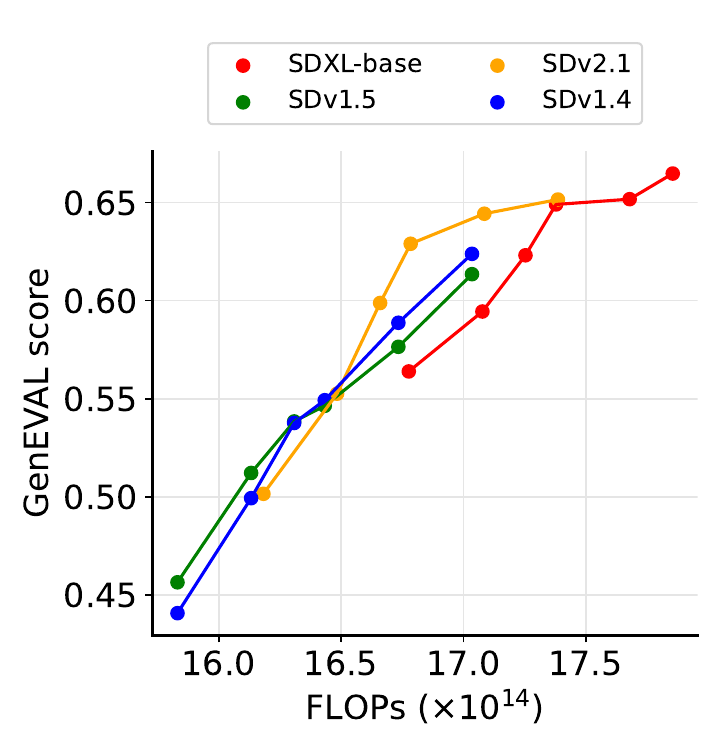}
    \caption{\textbf{With \gls{method} small models outperform bigger models with faster sampling and less compute}. We compare stable diffusion text-to-image models and their fine-tuned versions, \textsc{DPO} \citep{wallace2024diffusion} and DDPO \citep{black2023training}, with \gls{method}. We use the ImageReward model \citep{xu2024imagereward} as the reward function and generate samples from the base model without reward gradients \citep{bansal2023universal}. \textbf{We also note that \gls{method} with $k=2$ outperforms the fine-tuned models}. \textit{Left}: The GenEval prompt fidelity scores \citep{ghosh2024geneval} of the base models with \gls{method} and the fine-tuned models \citep{wallace2024diffusion}. Without \textit{any training} SDv2.1 with \gls{method} ($k=3$)  outperforms a finetuned SDXL model with {fewer FLOPS and less sampling time}. \textit{Right:} FLOPs for each configuration are on the x-axis. }\label{fig:main}
\end{figure*}

\gls{method} works by (a) sampling multiple interacting diffusion processes, called \textit{particles}, (b) scoring these particles using functions called \textit{potentials}, and (c) resampling the particles based on their potentials at intermediate steps during generation (see \cref{fig:fk_graphic}). Potential functions are defined using intermediate rewards. Resampling with these intermediate rewards yields high-reward samples and eliminates lower reward particles.
We present several ways of defining these intermediate rewards and potentials and empirically demonstrate that these new choices improve on traditional choices \citep{wu2024practical,li2024derivativefreeguidancecontinuousdiscrete}. By expanding the set of choices, users can find potentials that are better suited for their tasks.
Remarkably, for a number of tasks, we see significant performance benefits for both image and text diffusion models with \gls{method} with as few as $k=4$ particles (see \cref{fig:main}). Additionally, we find that \gls{method} smaller diffusion models outperforms larger models, and their fine-tuned versions, \textit{using less compute}. 

\begin{figure}[t]
    \centering
    \includegraphics[width=\linewidth]{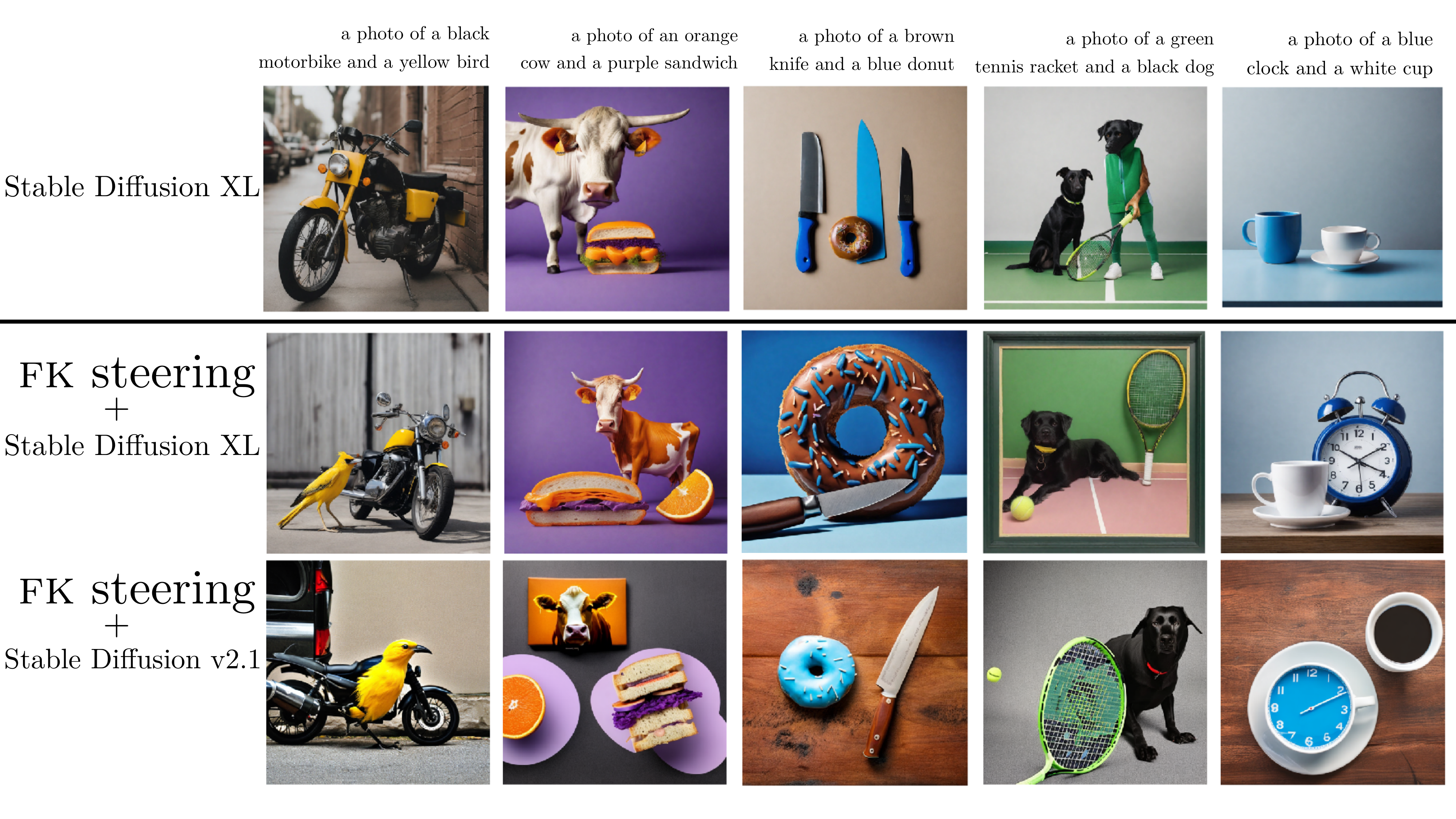}
    \caption{\textbf{\gls{method} improves prompt fidelity and sample quality.} In the \textit{first row}, we plot an independent sample from SDXL \citep{podell2023sdxl}. In the \textit{middle and bottom rows}, we plot the highest reward sample using \gls{method} with SDXL and SDv2.1 \citep{rombach2022high}, here we use $k=4$ particles. We use a human preference score, ImageReward \citep{xu2024imagereward}, as the reward function and generate samples from the base models, without any gradients from the reward.  We observe that \gls{method} both SDXL and SDv2.1 improves their prompt fidelity and sample quality compared to the first row. Prompts are selected from the GenEval benchmark prompt set.}
    \label{fig:geneval_prompt_figures}
\end{figure}

\paragraph{Contributions.}
In summary, our methodological contributions are the following:
\begin{itemize}[leftmargin=*]
    \item We propose Feynman-Kac diffusion steering as a framework for building particle-based approximations for sampling the tilted distribution $p_\theta(\mbx_0) \exp(\lambda r(\mbx_0))$. \gls{method} can steer both continuous and discrete state-space models with generic reward functions. We also show that \gls{method} can be used to steer conditional models.
    
    \item We show that particle-based sampling methods such as \gls{tds} \citep{wu2024practical} and \citet{li2024derivativefreeguidancecontinuousdiscrete}, are specific instances of \gls{fk} interacting particle systems. We demonstrate that \gls{method} enables new choices of potentials, samplers, and reward models that can improve performance across many tasks.
\end{itemize}

Empirically, we demonstrate that \gls{method}:
\begin{itemize}[leftmargin=*]
    \item Provides an alternative to fine-tuning when using sample quality rewards. We steer text-to-image latent diffusion models with an off-the-shelf reward model. \gls{method} with just $k=4$ particles outperforms fine-tuning on prompt fidelity and aesthetic quality, without making use of reward gradients. We also steer text diffusion models to generate higher quality samples that have more competitive linguistic acceptability and perplexity to those sampled from auto-regressive models.

    \item Enables smaller models (0.8B parameters) to outperform larger models (2.6B parameters) on prompt-fidelity 
    \citep{ghosh2024geneval}, using fewer FLOPs (see the right panel in \cref{fig:main}). Moreover, we show that using \gls{method} with fine-tuned models unlocks further performance benefits. 
    \item Outperforms fine-tuned models on prompt fidelity with just $k=2$ particles (see \cref{fig:main}).
    \item Provides a valuable method for generating samples with (rare) specified attributes. As an example, the method can be used to generate toxic language to support efforts in responsible AI \citep{zhao2024probabilisticinferencelanguagemodels}, including for red-teaming model behavior, and for fine-turning and steering generalist language models to recognize and reject toxic language input. On toxicity, without gradient guidance, \gls{method} can increase the toxicity rate of a text diffusion model from $0.3\%$ to $64.7\%$ with $k=8$ particles, and outperforms both gradient guidance and best-of-$n$. 
\end{itemize}
Overall, in \textit{all} settings we consider, \gls{method} \textit{always} improves performance, highlighting the benefits of inference-time scaling and steering of diffusion models.

\section{Related Work}\label{sec:related-work}
Current approaches to generate samples from the tilted distribution $p_\theta(\mbx_0) \exp(\lambda r(\mbx_0))$ can be categorized into two types: (1) fine-tuning  \citep{black2023training,domingo2024adjoint,fan2024reinforcement}, and (2) inference-time steering approaches, such as universal guidance \citep{song2020score,bansal2023universal} and particle-based approaches such as best-of-$n$ sampling and \gls{smc} \citep{wu2024practical}. 

\paragraph{Fine-tuning.} Recent work \citep{black2023training,xu2024imagereward} proposes fine-tuning a diffusion model $q_\text{finetune}$ to maximize the reward $\E_{q_\text{finetune}(\mbx_0)}r(\mbx_0)$ without a \gls{kl} penalty. \citet{fan2024reinforcement,domingo2024adjoint} propose \gls{kl}-regularized fine-tuning, and more recently \citet{wallace2024diffusion} proposes direct preference optimization for diffusion models. 
However, fine-tuning requires both allocating training resources and coupling a model to a specific reward function. Moreover, we show \gls{fk-ips}, with just $k=3$ particles, outperforms fine-tuning in several settings without requiring any kind of training.

\paragraph{Inference-time steering.} For continuous-valued $\mbx_t$ and differentiable rewards, one can employ gradient-based steering methods such as classifier guidance \citep{song2020score}, or more generally universal guidance \citep{bansal2023universal}, to define the \textit{twisted} score, $s_\theta(\mbx_t, t) + \nabla_{\mbx_t} r(\mbx_t)$, where $s_\theta$ is the marginal score. However, the use of gradients limits steering to differentiable rewards and continuous-valued diffusion models.

\gls{method} builds on top of recent works that sample from Feynman-Kac path distributions for conditional sampling with diffusion models, either using particle-based sampling \citep{trippe2022diffusion,wu2024practical,cardoso2023monte,dou2024diffusion,zhao2024conditional} 
or gradient-based sampling \citep{chung2022diffusion,janati2024divide}.

In \cref{sec:existing_work}, we show how \gls{tds} \citep{wu2024practical} and \gls{svdd} \citep{li2024derivativefreeguidancecontinuousdiscrete} are examples of \gls{fk} interacting particle systems \citep{moral2004feynman}. Our experiments demonstrate that generalizing beyond these methods to different choices of potentials, rewards, and samplers provide several improvements, such as higher reward samples. 

\gls{tds} \citep{wu2024practical} uses twisted \gls{smc} (See Section 3 in \citet{naesseth2019elements}) for conditional sampling by targeting $p_\text{target}(\mbx_0 \mid y) \propto p_\theta(\mbx_0) p(y \mid \mbx_0)$. The proposal generator for \gls{tds} uses classifier-guidance, restricting guidance to continuous state diffusion models and differentiable reward functions. In contrast, \gls{method} enables guidance with reward functions beyond differentiable likelihoods and generalizes to discrete state-spaces, including for text diffusion models.  

More recently, \citet{li2024derivativefreeguidancecontinuousdiscrete} propose \gls{svdd}, a derivative-free approach to guiding diffusion models for \textit{reward maximization}, by targeting the distribution the limit of $\lim_{\lambda \rightarrow \infty} \frac{1}{\mbZ} p_\theta(\mbx_0)\exp(\lambda r(\mbx_0))$. If there is a single sample $\mbx^*_0$ in the support of $p_\theta$ that maximizes $r$, then this distribution collapses to a point mass on $\mbx^*_0$. \gls{svdd} uses nested importance sampling for sampling, see algorithm 5 in \citet{naesseth2019elements}. At each time step, \gls{svdd} samples $k$ states $\mbx_t^i$ from a diffusion model and selects \textit{one state} $\mbx_t^{i*}$ with the highest reward and makes $k$ copies of that state, therefore reducing diversity of samples $\mbx_0^i$.

\begin{figure*}[h]
    \centering
    \includegraphics[width=0.80\linewidth]{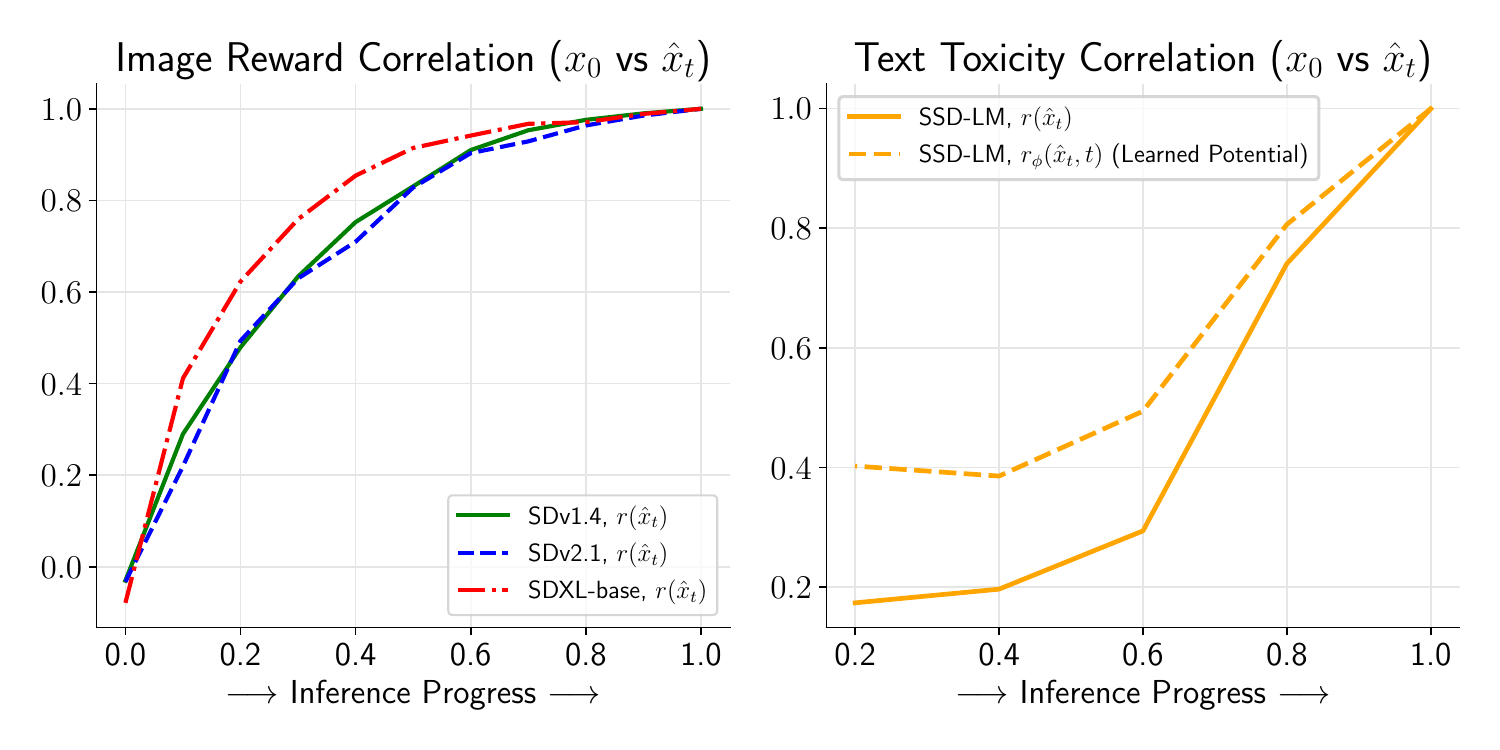}
    \caption{\textbf{Correlation between $r_\phi(\mbx_t)$ and final state $r(\mbx_0)$}: \textit{Left:} Correlations between $r(\mbx_{0})$ and $r(\mbx_0 = \widehat{\mbx}_t)$ for several text-to-image diffusion models, where $r$ is a human preference score \citep{xu2024imagereward}.  \textit{Right}: Correlation with an off-the-shelf text toxicity classifier $r(\mbx_0 = \widehat{\mbx}_t)$ and learned rewards $r_{\phi}(\mbx_t)$ on SSD-LM \citep{han2023ssdlmsemiautoregressivesimplexbaseddiffusion}, a text diffusion model. }
    \label{fig:reward-corr}
\end{figure*}

\paragraph{Text Generation and SMC.}  The sampling procedure for traditional autoregressive language models poses a challenge for \gls{smc} approaches, since \gls{smc} typically requires estimating the reward of the full sequence given its prefix. \cite{lew2023sequential} address this challenge by limiting to rewards calculated within a fixed look-ahead window. In contrast, \citet{zhao2024probabilisticinferencelanguagemodels} learn intermediate twisting potentials to marginalize over the remaining elements of a particular partial sequence. In our work, we demonstrate that intermediate estimates from diffusion models, for attributes like toxicity, can be used even with off-the-shelf reward models by evaluating the reward models on intermediate denoised estimates (see \cref{fig:reward-corr}).

\section{Feynman-Kac Steering of diffusion models}\label{sec:methods}
In this section, we present details of the Feynman-Kac steering (\gls{method}) for inference-time steering of diffusion models. 

\subsection{Diffusion Models}
\Glspl{dbgm} \citep{sohl2015deep} are processes that are learned by reversing a forward noising process, $q(\mbx_t)$. The noising process takes data $x \sim \qdata$ and produces a noisy state $\mbx_t \sim q(\mbx_t \mid \mbx_0=x)$ such that at a terminal-time $T$ we have $q(\mbx_T) = \pi_{\text{prior}}$, where $\pi_{\text{prior}}$ is the model prior. The noising process can be
defined as a continuous-time Markov process \citep{song2020score,kingma2021variational,singhal2023diffuse,lipman2022flow, singhal2024score} or discrete-time Markov chain  \citep{sohl2015deep,austin2021structured,sahoo2024simple,shi2024simplified,campbell2022continuous}. For exposition, we focus on discrete time diffusions though the techniques are applicable to continuous time diffusions as noted below. A discrete time diffusion has a model given a context $\mbc$
\begin{align}\label{eq:disc_time_diff}
    \text{Discrete-time:}  \qquad & p_\theta(\mbx_T, \dots, \mbx_0 \g \mbc) = \prior(\mbx_T) \prod_{t \in [T-1, 0]} p_\theta(\mbx_t \mid \mbx_{t+1}, \mbc),
\end{align}
which involves iteratively sampling a \textit{path} $(\mbx_T, \mbx_{T-1}, \dots, \mbx_0)$, where $\mbx_0$ is the model sample. The model $p_\theta$ can be trained by maximizing a lower bound on the model log-likelihood $\log p_\theta(\mbx_0 = x)$.

Most uses of generative models require samples with user-specified properties. In the next section, we describe a generic way to steer diffusion models towards such samples.

\subsection{Steering Diffusion Models}
One way to steer diffusion models is to encode user preferences in a reward model $r(\mbx_0)$ and sample from a distribution that tilts the diffusion model's generations $p_\theta(\mbx_0)$ towards an exponential of the reward function $r(\mbx_0)$:
\begin{align}\label{eq:p_target}
    \ptarget(\mbx_0 \g \mbc) &= \frac{1}{\mbZ} p_\theta(\mbx_0 \g \mbc) \exp \left(\lambda r(\mbx_0, \mbc)\right).
\end{align}
The reward $r(\mbx_0)$ can correspond to various objectives, including human preference scores \citep{xu2024imagereward,wu2023human},
vision-language models \citep{liu2024visual} that measure the prompt fidelity for text-to-image models, or likelihoods $p(y \mid \mbx_0)$ of a particular attribute $y$.

One way to sample from the target distribution in \cref{eq:p_target} is to generate $k$ \textit{particles} $\{\mbx_0^i\} \sim p_\theta(\mbx_0 \mid \mbc)$ and then resample the particles based on the scores $\exp(\lambda( r(\mbx_0^i, \mbc)))$. This procedure is known as importance sampling \citep{owen2000safe}. However, the target distribution favors samples that have higher reward, which may be rare under the model $p_\theta$. This suggests the use of simulation methods that better tilt towards rare events.

One broad class of rare-event simulation methods are \gls{fk-ips} approaches \citep{moral2004feynman,hairer2014improved} that use functions called \textit{potentials} to tilt the transition kernels of the diffusion process to push samples towards paths that lead to higher rewards. 
In the next section, we describe \gls{method}, a general framework for inference-time steering of diffusion models using \gls{fk-ips}.

\subsection{Feynman-Kac diffusion steering}
We use \gls{fk-ips} to produce paths  $(\mbx_T, \mbx_{T-1}, \dots, \mbx_0)$ with high-reward $\mbx_0$ samples. 
\gls{fk-ips} defines a sequence of distributions $\pfk{t}(\mbx_T, \mbx_{T-1}, \dots, \mbx_t)$ by tilting the base distribution $p_\theta(\mbx_T, \mbx_{T-1}, \dots, \mbx_t)$ using potentials $G_t$ \citep{moral2004feynman,chopin2020introduction}. The sequence of distributions $\pfk{t}$ is built iteratively by tilting the transition kernels $p_\theta(\mbx_t \mid \mbx_{t+1})$  with a potential $G_t(\mbx_T, \mbx_{T-1}, \dots, \mbx_t)$. We start with $\pfk{T}(\mbx_T) \propto p_\theta(\mbx_T \mid \mbc) G_T(\mbx_T, \mbc)$ and then define the subsequent distributions as:
\begin{align}
    \pfk{t}(\mbx_T, \dots, \mbx_t \mid \mbc) &= \frac{1}{\mbZ_t} p_\theta(\mbx_T, \mbx_{T-1}, \dots, \mbx_t \mid \mbc) \left\{\prod_{s=T}^t G_t(\mbx_T, \mbx_{T-1}, \dots, \mbx_s, \mbc) \right\}
    \\
    &= \frac{1}{\mbZ_t} \underbrace{p_\theta(\mbx_t \mid \mbx_{t+1},  \mbc) G_t(\mbx_T, \dots, \mbx_t, \mbc)}_{\text{Tilted transition kernel}} \underbrace{p_\theta(\mbx_T, \dots, \mbx_{t+1} \mid \mbc) \prod_{s=T}^{t+1} G_s(\mbx_T, \dots, \mbx_s, \mbc)}_{ \propto \text{Previous } \pfk{t+1}} 
\end{align}
where $\mbZ_t = \E_{p_\theta}[\prod_{s=T}^t G_s]$ is the normalization constant. The potentials $G_t$ are selected to up-weight paths $(\mbx_T, \dots, \mbx_t)$ that will ultimately yield high-reward samples $\mbx_0$. We require that the product of the potentials $G_t$ matches the exponential tilt of $\ptarget$:
\begin{align}\label{eq:pot_requriement}
    \prod_{t = T}^{0} G_t(\mbx_T, \dots, \mbx_t, \mbc) = \exp\left(\lambda r(\mbx_0, \mbc)\right).
\end{align}
This choice ensures
that $\pfk{0}(\mbx_T, \dots, \mbx_0 \mid \mbc) \propto p_\theta(\mbx_T, \dots, \mbx_0 \g \mbc) \exp(\lambda r(\mbx_0,\mbc))$; that is, sampling from $\mbx_0 \sim \pfk{0}$ samples from the target that tilts towards higher rewards. Potential functions that satisfy this constraint are not unique. 

\paragraph{Sampling from $\pfk{0}$.} 

Direct sampling from $\pfk{0}$ is intractable. However, targeting the intermediate distributions $\pfk{t}$ supports sampling of the distribution $\pfk{0}$ through the use of particle-based methods, such as \gls{smc} \citep{moral2004feynman,doucet2018sequential}, nested \textsc{is} (see alg. 5 in \citet{naesseth2019elements}), \gls{dmc} \citep{hairer2014improved}. \gls{smc} generates $k$ particles using a proposal generator $\tau(\mbx_t \mid \mbx_{t+1}, \dots, \mbx_T, \mbc)$ and at each transition step $\mbx_{t + 1} \rightarrow \mbx_t$ scores the particles using the potential and the transition kernel importance weights: 
\begin{align}    
    G^i_t &= \frac{\pfk{t}(\mbx_T, \dots, \mbx_{t+1}, \mbx_t \mid \mbc)}{\pfk{t+1}(\mbx_T, \dots, \mbx_{t+1} \mid \mbc) \tau(\mbx_t \mid \mbx_{t+1}, \dots, \mbx_T \mid \mbc)} \\ &= G_t(\mbx^i_T, \dots, \mbx_{t+1}^i, \mbx_t^i, \mbc) \frac{p_\theta(\mbx^i_t \mid \mbx^i_{t+1}, \mbc )}{\tau(\mbx^i_t \mid \mbx^i_{t+1}, \dots, \mbx^i_{T}, \mbc)} .
\end{align}
The particles $\mbx_t^i$ are then resampled
based on the scores $G_t^i$. 
See \cref{fig:fk_graphic} for a visualization of the method and \cref{alg:fk_sampling} for details. 
Particle-based approximations are consistent, that is as $k \rightarrow \infty$, the empirical distribution over the particles $\mbx_0^i$ converges to $p_\text{target}(\mbx_0)$, see theorem 3.19 in \citet{del2000branching}. Next, we discuss different choices of the potentials $G_t$ and proposal generators $\tau$.

\begin{algorithm}[t!]
\begin{algorithmic}
 \STATE {\bfseries Input:} Diffusion model $p_\theta(\mbx_{0:T} \mid \mbc)$, reward model $r(\mbx_0, \mbc)$, proposal generator $\tau(\mbx_t \mid \mbx_{t+1}, \mbc)$, potentials $G_t$, intermediate rewards $r_\phi(\mbx_t, \mbc)$, number of particles $k$    
     \STATE {\bfseries Returns:} Samples $\{\mbx^i_0\}_{i=1}^k$ 
     \STATE Sample $\mbx^{i}_T \sim \tau(\mbx_T \mid \mbc)$ for $i \in [K]$ 
     \STATE Score , $G_T^i = G_T(\mbx_T^i, \mbc)$ for $i \in [K]$ 
    \FOR{$t \in \{T, \dots, 1\}$}
        \STATE \textbf{Resample:} Sample $k$ indices $a^{i}_t \sim \text{Multinomial}(\mbx^i_t, G^i_t$) and let $\mbx_t^i = \mbx_{t}^{a_i}$ for $i \in [K]$
        \STATE \textbf{Propose:} Sample $\mbx_{t-1}^{i} \sim \tau(\mbx_{t-1} \mid \mbx^{i}_{t}, \mbc)$ for $i \in [K]$
        \STATE \textbf{Re-weight:} Compute weight $G_{t-1}^i$ for $i \in [K]$:
        \vspace{-0.3cm}
        \begin{align*}
         G^i_{t-1} = \frac{p_\theta(\mbx^i_{t-1} \mid \mbx^i_{t}, \mbc)}{\tau(\mbx^i_{t-1} \mid \mbx^i_{t}, \mbc)} G_{t-1}(\mbx^i_T, \dots, \mbx^i_{t-1}, \mbc)  
        \end{align*}
        \vspace{-0.5cm}
    \ENDFOR
    \STATE{ \bfseries Output:} return samples $\{\mbx_0^i\}$ 
\end{algorithmic}
\vskip -0.05in
\caption{Feynman-Kac Diffusion Steering}\label{alg:fk_sampling} 
\end{algorithm}

\paragraph{Choosing the proposal generator $\tau$.}
For the proposal generator $\tau$, the simplest choice is to sample from the diffusion model's transition kernel $p_\theta(\mbx_t \mid \mbx_{t+1},\mbc)$. Alternatively, another choice is a transition kernel that tilts towards samples with high rewards, such as gradient-based guidance. We discuss choices in \cref{sec:twisted_transition}.

\paragraph{Choosing the potential $G_t$.} One choice of potentials is $G_t = 1$ for $t \geq 1$ and $G_0 = \exp(\lambda(r(\mbx_0, \mbc)))$, which leads to importance sampling. However, importance sampling can require many particles to find high rewards \citep{chatterjee2018sample} and does not take into account how a particle is likely to score during the generation process. To up-weight paths that yield high-reward samples, \gls{method} uses potentials that score particles using \textit{intermediate rewards} $r_\phi(\mbx_t, \mbc)$:
\begin{itemize}[leftmargin=*]
    \item \textsc{Difference}: $G_t(\mbx_t, \mbx_{t+1}, \mbc) = \exp(\lambda(r_\phi(\mbx_t, \mbc) - r_\phi(\mbx_{t+1}, \mbc)))$ and $G_T = 1$, similar to \citep{wu2024practical}, prefers particles that have increasing rewards.
    \item \textsc{Max}: $G_t(\mbx_T, \dots, \mbx_t, \mbc) = \exp( \lambda \max_{s =t}^{T} r_\phi(\mbx_s, \mbc))$ and $G_0 = \exp(\lambda r(\mbx_0, \mbc)) (\prod_{t = 1}^T G_t)^{-1}$ prefers particles that have the highest rewards.
    \item \textsc{Sum}: $G_t(\mbx_T, \dots, \mbx_t) = \exp( \lambda \sum_{s =t}^{T} r_\phi(\mbx_s, \mbc))$ and $G_0 = \exp(\lambda r(\mbx_0, \mbc)) (\prod_{t = 1}^T G_t)^{-1}$ selects particles that have the highest accumulated rewards.
\end{itemize}
Any choice of potentials that satisfies \cref{eq:pot_requriement} produce a consistent approximation of $\ptarget(\mbx_0)$.
However, the number of particles required to produce high-reward samples depends on the choice of the potential. For instance, if the reward $r(\mbx_0)$ is bounded, then using the potential $\exp(\lambda(r_\phi(\mbx_t, \mbc) - r_\phi(\mbx_{t+1}, \mbc)))$ assigns low scores to particles that reach the maximum reward early in generation. In this setting, alternatives like the \textsc{Max} potential may offer benefits.

\textit{Interval Resampling.} For a typical diffusion process, the change between $\mbx_t$ and $\mbx_{t+1}$ is not substantial. Therefore, we can select potentials $G_t$ such that we only resample at a few steps. We define a \textit{resampling} schedule $R = \{t_r, \dots, t_1\}$, where $t_1 = 0$. For $t \not\in R$, $G_t = 1$ and for $t_j \in R$, $G_t$ is equal to a non-uniform potential, such as the max potential.
This type of interval resampling encourages exploration and reduces sampling time and compute requirements.

\paragraph{Choosing intermediate rewards $r_\phi(\mbx_t, \mbc)$.}\label{sec:intermediate_rewards}  %
Any choice of the intermediate rewards $r_{\phi}(\mbx_t, \mbc)$ yields consistent particle-based approximations. In this section, we discuss some choices of $r_\phi$.

The ideal rewards for the intermediate state $\mbx_t$ would require knowing the distribution of rewards of $\mbx_0$ generated from the state $\mbx_t$: $p_\theta(r(\mbx_0) \g \mbx_t, \mbc)$. With this distribution, reward functions $r_\phi$ could be chosen to yield high expected reward when reward variance is low, or based on the 10th percentile of the reward distribution to ensure good worst-case quality. Producing this distribution of rewards requires training with samples from the model. We will describe a few options that have different trade-offs in compute versus knowledge provided about the rewards for the sample $\mbx_0$.

\begin{itemize}[leftmargin=*]
    \item \textbf{Rewards at expected $\mbx_0$.} Similar to \citet{song2020score,bansal2023universal,wu2024practical,li2024derivativefreeguidancecontinuousdiscrete}, intermediate rewards can be defined by evaluating an off-the-shelf reward function at the diffusion model's approximation of the expected sample $\mbx_0$ given the current state $\mbx_t$ and context $\mbc$:  $\widehat{\mbx}_{t} \approx \E_{p_\theta(\mbx_0 \mid \mbx_t)}[\mbx_0 \mid \mbx_t]$. With this choice, the intermediate rewards are $r_\phi(\mbx_t,\mbc) = r(\mbx_0 = \widehat{\mbx}_{t},\mbc)$.

    \item \textbf{Many-sample $r_\phi$.} Diffusion models provide a means to sample $p_\theta(\mbx_0 \mid \mbx_t, \mbc)$. During inference, for each particle $\mbx_t^i$, we sample $N$ samples $\mbx_0^{i, j} \sim p_\theta(\mbx_0 \mid \mbx_t^i, \mbc)$ and then use  $r_\phi(\mbx_t^i, \mbc) = \log \frac{1}{N} \sum_{j=1}^N \exp(r(\mbx_0^{i, j}, \mbc))$
    to summarize the empirical distribution of rewards.

    \item \textbf{Learned $r_\phi$.} When sampling from $p_\theta(\mbx_0 \mid \mbx_t, \mbc)$ is expensive, we can leverage the fact that $p_\theta$ is trained to approximate the inference process $q$ \citep{sohl2015deep,song2020score}. Because of this approximation, data samples can be used to train intermediate reward models $r_\phi$. For instance, when $r(\mbx_0)$ is a classifier $p_\theta(y \mid \mbx_0)$, then similar to \citet{nichol2021glide} we can train a classifier $p_\phi(y \mid \mbx_t)$. For more general rewards, we can use:
    \begin{align}\label{eq:learned_r_phi}
        \argmin_{\phi} \E_{t \sim U[0,T]} \E_{\qdata(\mbx_0 \mid \mbc) q(\mbx_t \mid \mbx_0)} \norm{a_\phi(\mbx_t, \mbc) - \exp(r(\mbx_0, \mbc))}_2^2
    \end{align}
    and define $r_\phi = \log a_\phi$. When $p_\theta = q$, these objectives learn the reward, $r_\phi = \log \E_{p_\theta(\mbx_0 \mid \mbx_t, \mbc)}[\exp(r(\mbx_0, \mbc))]$. This choice of reward is used to define the potential $G_t$ that leads to the local transitions that minimize the variance of the potential at each step (see theorem 10.1 in \citep{chopin2020introduction}). 
\end{itemize}

\paragraph{Continuous-time diffusions.}  While the presentation above is for discrete-time models, we note that \gls{method} can also be used to steer continuous-time diffusion models \citep{song2020score, singhal2023diffuse}. Sampling from continuous-time diffusion models is done using numerical methods such as Euler-Maruyama \citep{sarkka2019applied}. Numerical sampling methods involve defining a discretized grid $\{1, 1 - \Delta t, \dots, 0\}$ and then sampling from the transition kernel $p_\theta(\mbx_t \mid \mbx_{t + \Delta}, \mbc)$. Therefore, similar to discrete-time models, we can apply \gls{method} by tilting the transition kernels with potentials $G_t(\mbx_1, \mbx_{1 - \Delta t}, \dots, \mbx_t)$ for $t \in \{0, \Delta t, \dots, 1\}$.

\section{Experiments}
To evaluate the efficacy of \gls{method} we conduct the following experiments:

\begin{itemize}[leftmargin=*]
    \item \textbf{\gls{method} for sample quality}: This experiment steers text-to-image diffusion models and text diffusion models with off-the-shelf rewards that measure sample quality. 

    \begin{itemize}
        \item  In the text-to-image setting, we run \gls{method} with a human preference reward model, ImageReward \citep{xu2024imagereward}. We evaluate on the heldout {GenEval} benchmark, a popular prompt fidelity evaluation.\footnote{\url{https://github.com/djghosh13/geneval/tree/main/prompts}}
        \item For text diffusion models, we use  perplexity computed using GPT2 \citep{radford2019language}, a simple trigram language model \citep{Liu2024InfiniGram}, and linguistic acceptability classifier \citep{textattack}  for rewards.
    \end{itemize}

     \item \textbf{Studying potential choices in \gls{method}}: In this experiment, we explore the effect of using different choices of potentials on the rewards achieved by the samples $\mbx_0^i$. 
            
    \item \textbf{Studying different choices of intermediate rewards}: We examine the effectiveness of using different intermediate rewards with \gls{method} on control of two types of rare attribute:
    \begin{itemize}
        \item For text diffusion models, we consider control of text \textit{toxicity}, which occurs in  $1\%$ of base model samples. 
        \item For image diffusion models, we evaluate control of ImageNet class. There are $1000$ classes in the dataset. In this experiment, we also use gradient-based guidance to tilt the transition kernel.
    \end{itemize}    
\end{itemize}

\subsection{\gls{method} for sample quality}\label{sec:sample_quality}

\paragraph{Text-to-Image Diffusion Models.} This experiment uses the stable diffusion \citep{rombach2022high,podell2023sdxl} family of text-to-image diffusion models $p_\theta(\mbx_0 \mid \mbc)$, where $\mbc$ is the text prompt. These models cover a range of model architectures \citep{nichol2021improved,peebles2023scalable} and inference processes \citep{ho2020denoising,karras2022elucidating}, see \cref{tab:param_count} for parameter counts and timings.  As the reward we use the ImageReward human preference score model \citep{xu2024imagereward}. For the intermediate rewards, we evaluate the off-the-shelf reward model on the denoised state, $r_\phi(\mbx_t) = r(\mbx_0 = \widehat{\mbx}_t)$ where $\widehat{\mbx}_t$ is the model's approximation of $\E_{p_\theta}[\mbx_0 \mid \mbx_t]$.

For the proposal generator $\tau$, we use the base model itself. For sampling from the base model, we use classifier-free guidance \citep{ho2022classifier} with guidance scale set to $7.5$, the default choice from Hugging Face\footnote{See \url{https://huggingface.co/blog/stable_diffusion}}, alongside the \textsc{ddim} sampler \citep{song2020denoising} with $\eta=1$ and $T=100$ time-steps. 
We  use $\lambda = 10$, $k=4$ and the resampling schedule $[0, 20, 40, 60, 80]$ with the potential $G_t(\mbx_T, \dots, \mbx_t) = \exp(\lambda \max_{s=t}^T r_\phi(\mbx_s))$, where $t=0$ is the terminal step.

As a benchmark for \gls{method}, we compare against best-of-N (BoN). Additionally, we consider publicly available models fine-tuned for prompt alignment and aesthetic quality. We consider \textsc{dpo}\footnote{\url{https://huggingface.co/papers/2311.12908}} fine-tuned models, SD v1.5 and SDXL, \citep{rafailov2024direct,wallace2024diffusion} and an RL fine-tuned SD v1.4 \citep{black2023training}\footnote{\url{https://huggingface.co/kvablack/ddpo-alignment}} with a vision-language model, \textsc{llava} \citep{liu2024visual}, as the reward for prompt alignment. We also explore \gls{method} the fine-tuned models.

We measure prompt alignment with the GenEval benchmark\footnote{\url{https://github.com/djghosh13/geneval/tree/main/prompts}} \citep{ghosh2024geneval} and report ImageReward\footnote{\url{https://github.com/THUDM/ImageReward/tree/main/benchmark}} \citep{xu2024imagereward} and \textsc{hps} \citep{wu2023human} scores to measure aesthetic quality. For results with different sampling schedules and $\lambda$ values, see \cref{sec:t2i_appendix}.

\begin{figure}[t]
    \centering
        \includegraphics[width=1.0\linewidth]{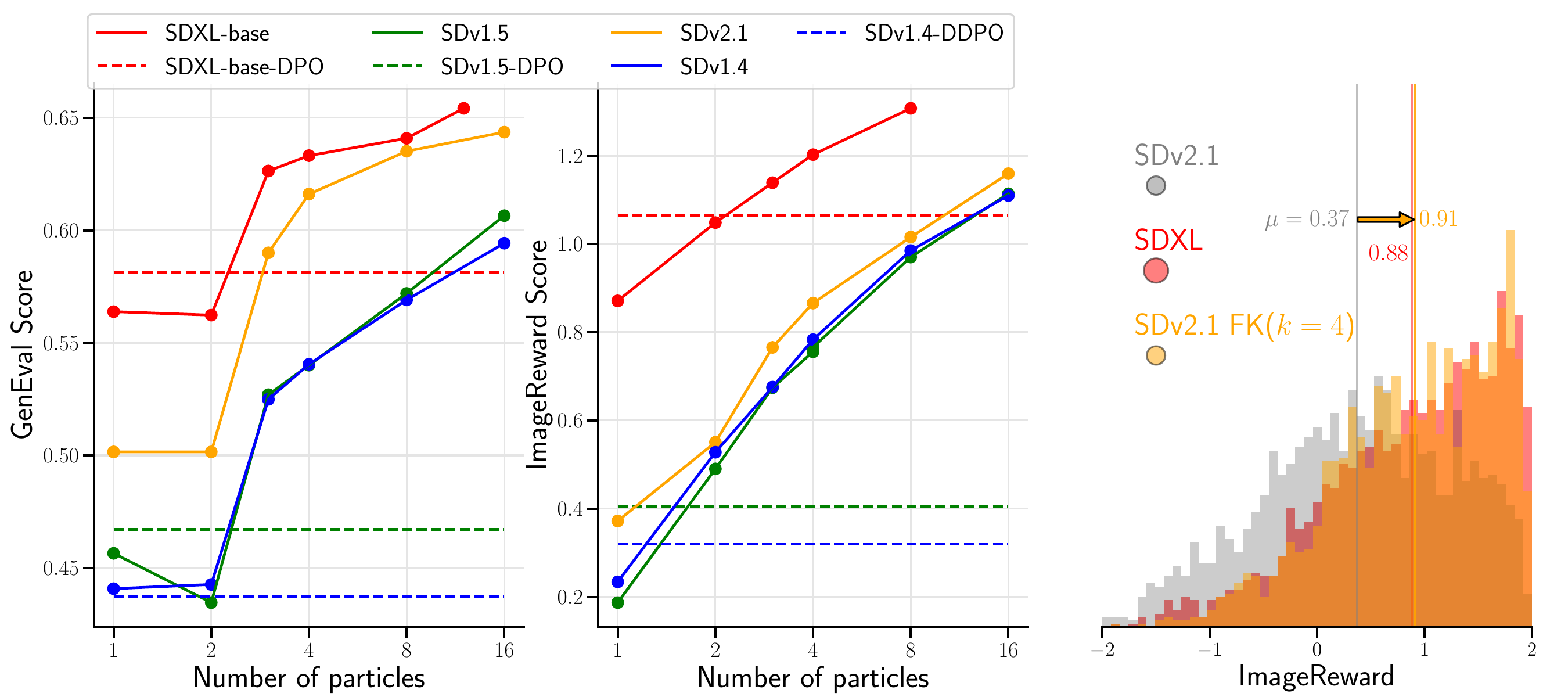}
    \caption{\textbf{Effect of scaling the number of particles.} \textit{Left}: \textsc{GenEval} scores for \gls{method} using \textsc{ImageReward}, average particle performance. Dashed lines indicate performance of fine-tuned baselines. \textit{Middle}: Corresponding \textsc{ImageReward} scores. \textit{Right}: Distribution of \textsc{ImageReward} scores for samples from SDv2.1 (0.8B) with and without \gls{method}, compared with SDXL (2.6B).}\label{fig:geneval_score_ir_hist} 
\end{figure}

In \Cref{table:ir_geneval_t2i},  we measure prompt 
fidelity and aesthetic quality of samples generated by \gls{method} with $k=4$. We measure prompt alignment by GenEVAL and aesthetic quality using ImageReward \citep{xu2024imagereward} and \textsc{hps} scores. In \cref{table:ir_geneval_t2i} we report the performance of the highest reward particle generated by \gls{method}, and in \cref{fig:geneval_score_ir_hist}, we report average particle performance. We observe:
\begin{itemize}[leftmargin=*]
    \item \textbf{\gls{method} smaller models outperforms larger models.} With $k=4$ \gls{method} SDv2.1 outperforms SDXL and its DPO \citep{wallace2024diffusion} fine-tuned version, on  GenEval scores and aesthetic quality with less sampling time: $11.5$s versus $9.1$s. See \cref{fig:geneval_prompt_figures} for samples.
    
    \item \textbf{\gls{method} the base model outperforms fine-tuning the base model.} \gls{method} with as few as $k=4$ particles can outperform fine-tuned models on both prompt fidelity and human preference alignment. Moreover, \Cref{fig:main} shows that the base model with just $k=2$ particles has a \textit{higher GenEval score} than the \textsc{dpo} and \textsc{ddpo} fine-tuned models.
    
    \item \textbf{\gls{method} further improves fine-tuned models.} In \cref{tab:fine-tuned_model_evals}, we show that steering fine-tuned models with \gls{method} improves performance even further. For instance, the GenEval score of SDXL-DPO increases from $0.58$ to $0.65$, and similarly for SDv1.5, the GenEval score increases from $0.46$ to $0.56$.

    \item \textbf{\gls{method} can improve gradient-guidance.} In \cref{tab:gradient_guidance}, we show that \gls{method} SDv1.5 with $k=4$ outperforms gradient-guidance using the ImageReward model, using less sampling time. We note that using gradient-guidance for the proposal generator can improve performance even further, at the cost of increased sampling time and more compute.
    
    \item \textbf{Effect of scaling the number of particles.} In \cref{fig:geneval_score_ir_hist}, we observe the effect of scaling the number of particles on prompt fidelity and human preference alignment metrics. We note that scaling the number of particles improves the prompt fidelity and human preference alignment scores of all particles for all models.    
    
\end{itemize}

\begin{table}[t]
\begin{center}
\begin{small}
\centering
\begin{tabular}{lcccccc}
\toprule
\textbf{Model} &  \textbf{Sampler}($\lambda$, $k$) & \textbf{GenEVAL$^*$} $\uparrow$ & \textbf{IR}\textsuperscript{\textdagger} $\uparrow$ & \textbf{HPS}\textsuperscript{\textdagger} $\uparrow$ \\
\midrule
SD v1.4  & $k=1$ &  0.4408 & 0.234 & 0.245  \\
SD v1.4 & BoN($k=4$)   & 0.5460 & 0.800 & 0.256 \\
SD v1.4-\textsc{ddpo} & $k=1$  & 0.4371 &  0.263   &  0.241 \\
\textbf{SD v1.4} & {FK($\lambda = 10$, $k=4$)}  & \textbf{0.5492} & \textbf{0.927} & \textbf{0.263}   \\
\midrule
\midrule
SD v1.5 & $k=1$ & 0.4483 &  0.187 & 0.245  \\
SD v1.5 & BoN($k=4$)  & 0.5239 &   0.737 & 0.265   \\
SD v1.5-\textsc{dpo} & $k=1$ & 0.4671 & 0.343  & 0.255  \\
\textbf{SD v1.5} & {FK($\lambda = 10$, $k=4$)}  & \textbf{0.5463} & \textbf{0.898} & \textbf{0.263}   \\
\midrule
\midrule
SD v2.1 & $k=1$ & 0.5104 & 0.372  & 0.253  \\
SD v2.1 & BoN($k=4$)  & 0.6172 & 0.888 &  0.263  \\
\textbf{SD v2.1} & {FK}($\lambda = 10$, $k=3$)   & {0.5987} & {0.864} & {0.265}  \\
\textbf{SD v2.1} & {FK}($\lambda = 10$, $k=4$)   & \textbf{0.6289} & \textbf{1.006} & \textbf{0.268} \\
\midrule
\midrule
SDXL & $k=1$ & 0.5571 & 0.871  & 0.289  \\
SDXL & BoN($k=4$) & 0.6347 & 1.236 &  {0.296} \\
SDXL-\textsc{dpo} & $k=1$  & 0.5811 & 0.859  & {0.296}   \\
\textbf{SDXL} & {FK($\lambda = 10$, $k=4$)}   & \textbf{0.6489} & \textbf{1.298} &  \textbf{0.302}   \\
\midrule
\bottomrule
\end{tabular}
\end{small}
\end{center}
\caption{\textbf{Effect of \gls{method} prompt fidelity and human preference scores}. 
GenEval scores $^*$ are computed using the GenEval prompts and ImageReward\textsuperscript{\textdagger} and ImageReward and HPS scores\textsuperscript{\textdagger} are computed on prompts provided by the ImageReward paper. As a baseline, we compare against Best-of-N with $4$ independent samples. Across all models, \gls{method} improves both prompt fidelity as well as human preference alignment scores, beating BoN and fine-tuning. Interestingly, we note that BoN outperforms fine-tuning as well, showing the benefits of inference-time scaling.}\label{table:ir_geneval_t2i}
\end{table}

\begin{table}[h]
    \centering    
    \begin{tabular}{lcccc}
    \toprule
       \textbf{Model} &  Sampler($\lambda$, $k$)  & \textbf{GenEval} $\uparrow$ & \textbf{IR} $\uparrow$ & \textbf{HPS}  $\uparrow$ \\    
        \midrule
        SD v1.5-\textsc{dpo} & $k=1$ & 0.4671 & 0.343   & 0.255  \\
        \textbf{SD v1.5-\textsc{dpo}} & {FK}($\lambda = 10, k=4$)& \textbf{0.5751} &  0.885 & \textbf{0.276} \\           
        \midrule
        SDXL-\textsc{dpo} & $k=1$ & 0.5811 & 0.859 & 0.296   \\
        \textbf{SDXL-\textsc{dpo}} &  {FK}($\lambda = 10, k=4$)& \textbf{0.6755} & \textbf{1.198}    & \textbf{0.317} \\                                
    \bottomrule        
    \end{tabular}
    \caption{\textbf{Steering fine-tuned models.} Here we sample from $q_\text{finetune}(\mbx_0 \mid \mbc) \exp(r(\mbx_0, \mbc))$ using \gls{method}, where $q_\text{finetune}$ are \textsc{dpo} fine-tuned models \citep{wallace2024diffusion}. Note that all metrics are improved with \gls{method}. }
    \label{tab:fine-tuned_model_evals}
\end{table}

\begin{table}
    \centering    
    \begin{tabular}{llcccc}
    \toprule
       \textbf{Model}  & \textbf{Sampler} & \textbf{GenEval} & \textbf{IR}  & \textbf{HPS} & \textbf{Time} \\    
    \midrule       
        SDv1.5 & $k=1$ & 0.44 & 0.187  & 0.245 & 2.4s \\
        SDv1.5  & $\nabla$($k=1$) & 0.45 & 0.668  & 0.245 & 20s \\
        SDv1.5 & \textsc{fk}($k=4$) & 0.54 & 0.898  & 0.263 & 8.1s \\
        SDv1.5 &  \textsc{fk}($\nabla$, $k=4$) & 0.56 & 1.290  & 0.268 & 55s  \\    
    \bottomrule        
    \end{tabular}   
    \caption{\textbf{Comparison against gradient guidance.} Here we note that \gls{method} with the model as the proposal generator outperforms gradient guidance, with faster sampling. We also note that \gls{method} can benefit from gradient guidance, albeit at the cost of more compute and sampling time.}\label{tab:gradient_guidance}
    \vspace{-0.2cm}
\end{table}

\paragraph{Text Diffusion Models.} Next, we investigate steering with \gls{method} to improve the sample quality of text diffusion models, which generally underperform traditional autoregressive models on fluency metrics like perplexity \citep{li2022diffusionlmimprovescontrollabletext, gulrajani2023likelihoodbaseddiffusionlanguagemodels, horvitz2024paraguideguideddiffusionparaphrasers, sahoo2024simple}. We consider three reward functions for improving text quality: perplexity computed with a simple \textit{trigram language model}\footnote{We compute trigram probabilities using  $\infty$-gram \citep{Liu2024InfiniGram}.}, a classifier \citep{textattack}\footnote{\url{https://huggingface.co/textattack/roberta-base-CoLA}} trained on the Corpus of Linguistic Acceptability (CoLA) dataset \citep{cola}, and perplexity computed by GPT2 \citep{Radford2019LanguageMA}. For all choices of reward models, we define the intermediate rewards using $r_{\phi}(\mbx_t) = r(\mbx_0 = \widehat{\mbx}_t)$ and use the potential $G_t = \exp(\lambda (r_\phi(\mbx_t) - r_\phi(\mbx_{t+1})))$

We consider two base text diffusion models: SSD-LM \citep{han2023ssdlmsemiautoregressivesimplexbaseddiffusion} and MDLM \citep{sahoo2024simple} and use these models as the proposal generator $\tau$. SSD-LM is a continuous space diffusion model trained on noised word logits, while MDLM is a discrete diffusion model. 

For \gls{method}, we resample $50$ times during the inference process (every $10$ steps for SSD-LM and every $20$ or MDLM). For all models we use $\lambda=10.0$, and return the highest scoring sample at $t=0$. Following \citet{han2023ssdlmsemiautoregressivesimplexbaseddiffusion}, we generate $20$ continuations of length $50$ using their $15$ controllable generation prompts. In addition to \gls{method}, we evaluate base model performance and best-of-$n$. 
As a baseline to compare quality improvements from scaling inference steps \citep{gong2023diffuseqsequencesequencetext,sahoo2024simple}, we also include results for SSD-LM with $T=5000$ (versus $T=500$).  We additionally include results for GPT2-Medium \citep{Radford2019LanguageMA} as an auto-regressive baseline.
We evaluate all methods using perplexity computed with GPT2-XL \citep{Radford2019LanguageMA}, CoLA acceptability, and distinct uni/bi/trigrams (\textit{Dist-1/2/3}) \citep{han2023ssdlmsemiautoregressivesimplexbaseddiffusion,li-etal-2016-diversity}. Additional details on our text experiments are included in \ref{sec:text_q_appendix}.

\begin{table*}[t]
\begin{center}
\begin{small}
\begin{tabular}{lcccc}
\textbf{Model + Sampler($r$)} & $k$ & \textbf{PPL (GPT-XL)} $\downarrow$ & \textbf{CoLA} $\uparrow$ & \textbf{Dist-1/2/3} $\uparrow$ \\
\midrule

{{GPT2-medium}} & $1$ & 14.1 & 87.6 & 54/89/94 \\

\midrule

{{SSD-LM}} & $1$ & 23.2 & 68.3 & 46/83/92 \\
{{SSD-LM}}$_{T\times10}$  & $1$ & 18.8 & 76.6 & 46/81/90 \\

\midrule

\textbf{{{FK(GPT2)} }}  & $4$ &  \textbf{11.0} & 80.0 & 40/73/86 \\
\textbf{{{FK(Trigram)} }}  & $4$ &  14.1 & 77.4 & 41/76/88 \\

\textbf{{{FK(CoLA)}}}   & $4$ & 17.4 & \underline{95.7} & 45/80/90 \\

{{BoN(GPT2)}}  & $4$ &  13.6 & 75.6 & 42/77/88 \\

{{BoN(Trigram)}}  & $4$ &  15.9 & 71.9 & 43/78/89 \\

BoN(CoLA)  & $4$ & 19.2 & 93.8 & 46/81/91 \\

\midrule

{{BoN(GPT2)}}  & $8$ &  \underline{11.2} & 80.3 & 41/74/86 \\

{{BoN(Trigram)}}  & $8$ & 13.9 & 76.8 & 42/77/89 \\

BoN(CoLA)  & $8$  & 18.4 & \textbf{97.2} & 46/82/91  \\

\midrule

{{MDLM}}  & $1$ & 85.3 & 28.9 & 57/91/94 \\
\midrule
\textbf{{{FK(GPT2)}}}   & $4$   &  49.0 & 39.8 & 52/86/92\\
\textbf{{{FK(Trigram)}}}   & $4$  & \textbf{40.3} & 37.0 & 50/87/93  \\
\textbf{{{FK(CoLA)}}}   & $4$ & 73.6 & \underline{69.8} & 59/88/91 \\

{{BoN(GPT2)}}  & $4$  &  55.5 & 32.9 & 54/88/93 \\

{{BoN(Trigram)}}  & $4$ & 52.1 & 30.1 & 53/89/93 \\

{{BoN(CoLA)}}   & $4$ & 71.4 & 59.4 & 57/90/94 \\
\midrule

{{BoN(GPT2)}}  & $8$  &  46.9 & 37.2 & 53/87/92 \\

{{BoN(Trigram)}}  & $8$ & \underline{45.9} & 35.4 & 52/88/93 \\
{{BoN(CoLA)}}   & $8$ & 68.2 & \textbf{73.1} & 58/91/94 \\
\midrule

\bottomrule
\end{tabular}
\end{small}
\end{center}
\caption{ \textbf{Text sample quality results metrics.} We sample texts of length $50$ from all models and score perplexity with GPT2-XL and CoLA acceptability. Results are averaged over three seeds. Both \textsc{SSD-LM} and \textsc{GPT}-medium have $355$ million parameters. \textsc{MDLM} is a smaller model with $170$ million parameters.
}\label{tab:hq-text}\vspace{-0.5cm}
\end{table*}

Table \ref{tab:hq-text} contains our text sample quality evaluation results. We observe:
\begin{itemize}[leftmargin=*]
    \item \textbf{\gls{method} significantly improves the perplexity and CoLA scores of both SSD-LM and MDLM}.  For all reward functions, \gls{method} ($k=4$) outperforms best-of-$n$ ($n=4$) on the corresponding target metric (perplexity or CoLA).  For MDLM, trigram steering dramatically improves perplexity ($37.2$ vs $79.2$), but is less effective at improving CoLA ($35.3$ vs $30.0$).
    \item \textbf{\gls{method} outperforms best-of-$n$.} For all experiments, \gls{method} outperforms best-of-$n$ when using the same number of particles. Notably,  \gls{method} SSD-LM outperforms best-of-$n$ with twice as many particles. We also note that \gls{method} on SSD-LM with $T=500$ improves on SSD-LM  with $10 \times$ inference steps ($T=5000$) for all fluency metrics.    
\end{itemize}

Overall, our results demonstrate that \gls{method} with off-the-shelf rewards can enable sampling lower-perplexity, more linguistically acceptable text from diffusion models.

\begin{table}[h]
    \centering    
    \begin{tabular}{lccccc}
    \toprule
       \textbf{Potential}  & $k$ & SDv1.4 & SDv1.5  & SDv2.1 & SDXL \\    
    \midrule       
        Max & $4$ & \textbf{0.540} & \textbf{0.540}  & \textbf{0.616} & \textbf{0.633} \\
        Sum & $4$ & 0.496 & 0.499  & 0.569 & 0.613 \\        
        Difference & $4$ & 0.525 & 0.526   & 0.578 & 0.603 \\        
        \midrule
        Max & $8$ & \textbf{0.569} & \textbf{0.561}   & \textbf{0.635} & \textbf{0.648} \\
        Sum & $8$ & 0.532 & 0.517   & 0.588 & 0.634 \\        
        Difference & $8$ & 0.566 & 0.553  & 0.615 & 0.640 \\        
    \bottomrule        
    \end{tabular}
    \caption{\textbf{Effect of different potentials on GenEval scores.} Here we plot the average GenEval prompt fidelity score, averaged over all particles. Using the max potential outperforms the difference potential and the sum potential.}\label{tab:generic_potentials}
\end{table}

\subsection{Studying different choices of potentials} 
In the previous section, we used two different choices of potentials: the max potential, $\exp(\lambda \max_{s\geq t} r_\phi(\mbx_s))$, for the text-to-image experiments and the difference potential, $\exp(\lambda(r_\phi(\mbx_t) - r_\phi(\mbx_{t+1})))$, for the text quality experiment. However, as discussed in \cref{sec:methods}, the choice of potential is not unique. 
This experiment studies different choices of the potential for steering text-to-image diffusion models. Similar to the previous section, we steer the stable diffusion text-to-image models with ImageReward, using the max, difference and the sum potential. For all potentials, we use  $\lambda=10$ with the $[0, 20, 40, 60, 80]$ interval sampling schedule. We also note that ImageReward is bounded between $[-2, 2]$.

In \cref{tab:generic_potentials}, for all models considered, the prompt fidelity scores using the max potential are higher compared to the difference and sum potentials. The sum potential is worse on the smaller models. When limiting to the difference and max potentials, the difference potential can assign low scores for particles that achieve the maximum reward of 2 early in generation (because no further reward increase can be made after reaching the maximum).
However, we observe that for the same value of $\lambda$ and same number of particles, the max potential can lead to lower particle diversity than the difference potential (see \cref{sec:samples} for samples). This is because resampling at intermediate steps with the max potential favors higher scoring particles more so than the difference potential.

\subsection{Studying different choices of rewards in \gls{method}}
In this experiment, we demonstrate the efficacy of \gls{method} for controllable generation, where we generate samples with attributes such as (a) toxic text for text diffusion models and (b) class-conditional image generation with $1000$ classes in the dataset. In these experiments, we also compare against classifier-guidance.

\begin{table*}
\begin{center}
\begin{small}
\begin{tabular}{lccccc}
\toprule
\textbf{Model + Sampler} & \textbf{Toxic} $\uparrow$ & \textbf{Toxic (Holdout)} $\uparrow$ & \textbf{PPL (GPT2-XL)} $\downarrow$ & \textbf{Dist-1/2/3} $\uparrow$ \\

{SSD-LM} & 0.4\% & 1.2\% & \textbf{23.2} & 46/83/92 \\
{SSD-LM} ($\nabla$ guidance) & 22.3\% & 22.6\% & 40.3 & 53/89/94  \\
MDLM & 0.3\% & 1.9\% & 85.3 & 57/91/94 \\
\midrule
\multicolumn{5}{l}{{SSD-LM (no gradients)}} \\
\midrule
{{BoN(4)}} & 1.6\% & 4.8\% & 21.9 & 46/81/91 \\
{{BoN(8)}} & 5.0\% & 8.1\% & 23.0 & 46/81/91 \\

\textbf{{{FK($k=4$)}}} & 8.4\% & 14.0\%  & 22.5 & 45/81/91 \\
\textbf{{{FK($k=4$, learned $r_\phi$)}}} & 15.2\% & 19.6\%  & 26.3 & 45/83/91 \\
\textbf{{{FK($k=8$)}}} & 25.0\% & 29.7\%  & 23.9 & 45/81/91 \\
\textbf{{{FK($k=8$, learned $r_\phi$)}}} & \textbf{39.0\%} & \textbf{38.0\%}  & 26.9  & 45/83/91 \\

\midrule
\multicolumn{5}{l}{{MDLM (discrete, no gradients)}} \\
\midrule

{{BoN(4)}} & 2.2\% & 6.7\% & 83.8 & 57/90/93\\
{{BoN(8)}} & 3.7\% & 10.8\% & 84.6 & 57/90/93  \\
\textbf{{{FK($k=4$)}}} & 23.0\% & 29.0\% & 81.0 & 56/90/93 \\
\textbf{{{FK($k=4$, many $r_\phi$)}}}  & 37.0\% & 40.2\% & 83.0 & 57/90/92 \\

\textbf{{{FK($k=8$)}}} & 53.4\% & 48.3\% & 74.3 & 56/89/92 \\
\textbf{{{FK($k=8$, many $r_\phi$)}}} & \textbf{64.7\%} & \textbf{51.7\%} & {82.9} & 57/89/92  \\
\bottomrule
\end{tabular}
\caption{\textbf{Toxicity results}. We evaluate the toxicity of the generated samples with (a) the classifier used for steering and (b) a separate holdout classifier, we also report GPT2-XL perplexity. Results are averaged over three seeds.}
\label{tab:text-diff-toxic-control}
\end{small}
\end{center}
\end{table*}

\paragraph{Controlling Text Toxicity.}
We consider the task of red-teaming \textit{toxicity}, a rare attribute identified in only $1\%$ of base SSD-LM samples and $0.3\%$ of MDLM samples. Toxicity presents an undesirable behavior for language models. Here, we examine whether \gls{method} enables examining rare but dangerous model behavior, a critical factor considered before deploying systems \citep{zhao2024probabilisticinferencelanguagemodels}. 
In this experiment, we run \gls{method} with the base text diffusion models with SSD-LM \citep{han2023ssdlmsemiautoregressivesimplexbaseddiffusion} and MDLM models. We use the base model as the proposal generator. As a baseline, we compare against gradient guidance for SSD-LM. For reward, we use a popular  toxicity classifier \citep{logacheva-etal-2022-paradetox}.\footnote{\url{https://huggingface.co/s-nlp/roberta_toxicity_classifier}}

In this experiment, we explore the effect of different choices of intermediate rewards:
\begin{itemize}[leftmargin=*]
    \item  For SSD-LM, we consider two choices: (1) the reward model evaluated at the denoised state and (2) the learned reward $r_\phi$, trained using real data samples.
    \item For MDLM, we use multiple samples $\mbx_{0}^{i, j} \sim p_\theta(\mbx_0^{i, j} \mid \mbx_t^i)$ to compute the reward $r_\phi = \log \frac{1}{N} \sum_{j=1}^N \exp(r(\mbx_0^{i, j}))$ with (1) $N=4$ samples and (2) $N=16$ samples.    
\end{itemize}

The sampling hyper-parameters and prompts are similar to \cref{sec:sample_quality}. In our evaluation, we also include results from an additional holdout toxicity classifier, trained on a multilingual mixture of toxicity datasets \citep{dementieva2024overview}.\footnote{\url{https://huggingface.co/textdetox/xlmr-large-toxicity-classifier}}  Details are included in \cref{sec:text_tox_appendix}.

In \Cref{tab:text-diff-toxic-control}, we observe the following:
\begin{itemize}[leftmargin=*]
    \item \textbf{Learned rewards improves controllability:} With $8$ particles, \gls{method} SSD-LM outperforms gradient guidance on fraction of generations labeled toxic ($29.7\%$ vs $22.6\%$) by the holdout classifier, and dramatically improves on perplexity ($23.9$ vs $40.3$).  Using improved intermediate rewards, learned from real data, improves this fraction even further to $39\%$. Additionally, in \cref{tab:reward_gradient_guided_toxicity}, we show that using learned reward gradient guidance with SS-LM with \gls{method} improves performance even further to $56.3\%$.
    
    \item \textbf{Using many-sample $r_\phi$ improves controllability:} \gls{method} MDLM with $k=8$ achieves an even higher fraction of text labeled toxic $48.3\%$. \gls{method} outperforms best-of-$n$ sampling with both $4,8$ particles. Using improved intermediate rewards for MDLM more samples for intermediate rewards improves performance even further to $51.7\%$ ($64.7\%$ on the guidance reward model).
    
\end{itemize}

\begin{table}[h]
    \centering    
    \begin{tabular}{lcccc}
    \toprule
       \textbf{Sampler} & $r_\phi(\mbx_t)$  & \textbf{Potential}   & $k$ & \textbf{$p(y \mid \mbx_0)$ Mean (Max)}  \\    
    \midrule       
        \gls{method} (\gls{tds} \citep{wu2024practical}) & $r(\mbx_0 = \widehat{\mbx}_t)$ & Diff. & 4 & 0.59 (0.72) \\  
        \gls{method} & $r(\mbx_0 = \widehat{\mbx}_t)$ & Max & 4 & 0.65 (0.70) \\       
        \midrule      
     \gls{method} & Learned & Diff.  & 4 & 0.88 (0.94) \\
     \gls{method} & Learned & Max  & 4 & 0.88 (0.96) \\
    \bottomrule        
    \end{tabular}
    \caption{\textbf{ImageNet class-conditional probabilities with different choices of rewards and potentials.} In this experiment, we explore the effect of two choices of rewards, learned and the reward evaluated at the denoised state \citep{wu2024practical}. We also explore the effect of different choices of potentials, the difference and the max potential. We observe that learning the reward improves performance significantly.}
    \label{tab:imagenet}
\end{table}

\paragraph{Class-Conditional Image Generation.} In this experiment, we steer a marginal diffusion model $p_\theta(\mbx_0)$ to produce samples from one of $1000$ different classes. Similar to \citet{wu2024practical}, the reward is $r(\mbx_0, y) = \log p_\theta(y \mid \mbx_0)$ and use gradient guidance for the proposal distribution.

We compare two potentials, the max potential and the difference potentials, along with two different reward models: one that uses the denoised state $r(\mbx_0 = \widehat{\mbx}_t, y)$ and one that is trained on noisy states $\mbx_t \sim q(\mbx_t \mid \mbx_0)$ where $\mbx_0 \sim \qdata$~\citep{nichol2021glide}. This experiment uses pre-trained marginal diffusion model and classifiers from \citet{nichol2021improved} and generate $256\times 256$ resolution images. 
In \cref{tab:imagenet}, we observe that learning $r_\phi$, for both gradient guidance and potential computation, provides significant improvements over the reward evaluated at the denoised state.

\paragraph{Better Rewards vs. More Particles.} In the previous section, we observed that for the same number of particles, using either learned rewards or multiple samples improves performance.  For instance, using $k=8$ with SSDLM without learning the intermediate rewards gets  a toxicity rate of $25\%$, higher than the learned reward's toxicity rate of $22.3\%$; using the learned rewards improves the toxicity rate to $39\%$. However, these rewards come with an added computational cost, either in training or more evaluations of the reward model, presenting a trade-off in terms of using extra computational resources either for more particles or for better intermediate rewards.

\section{Conclusion}
We introduce Feynman-Kac steering, a novel and efficient approach to inference-time steering of diffusion modeling. The method offers an extensible and scalable approach for improving diffusion models. \gls{method} flexibly steers diffusion models using Feynman-Kac interacting particle systems, a rare-event simulation technique. Our experiments demonstrate that \gls{method} can boost both the sample quality and controllability of image and text diffusion models, outperforming fine-tuning and other inference-time approaches. 

\gls{method} can be used in a ``plug-and-play" fashion to improve diffusion models on various downstream tasks. We observe that using (a) either the difference or the max potential with intermediate rewards defined at the denoised state and (b) the base diffusion model as the proposal generator improves performance significantly. For instance, we find \gls{method} smaller diffusion models, with off-the-shelf rewards, outperforms larger models, \textit{while using less compute}. However, we find that using learned rewards or rewards with many samples improves performance even further. Therefore, by validating different choices, such as potentials, rewards and samplers, a user can optimize performance for their task.

In our experiments, we find that scaling the number of particles is a natural mechanism for improving the performance of diffusion models. Notably, in our text-to-image experiments, \textit{even naive best-of-$n$ with $4$ particles outperforms fine-tuned models} and improves prompt fidelity and aesthetic quality. \gls{method} improves on best-of-$n$ by resampling using intermediate rewards during inference.
We explore several choices for these intermediate rewards, which present several trade-offs, including sample diversity versus high rewards and spending more compute for better intermediate rewards.

Promising future directions include exploring the value of varying the numbers of particles at inference time, either by having a variable budget or allocating compute adaptively. For applications like protein design, dynamically assigning greater numbers of particles to promising regions could enable generating a large number of diverse candidates. We also note that \gls{method} can also improve preference learning algorithms \citep{zhangpreference} by improving the sample generator.  Additionally, we believe it is crucial to better understand the limits of inference-time particle scaling and the corresponding compute-performance trade-offs.  

A limitation of \gls{method} and other inference scaling approaches is their reliance on the availability of strong reward functions. Therefore, advancing automated evaluation and reward modeling remains a critical area of research and can unlock further improvements for these methods.

\section*{Societal impact}
Controllable generation methods such as \gls{method} can be applied to align language models with human preferences, including to improve their personalization or safety.
Additionally, we show that \gls{method} can be used for automated red-teaming, which can inform model deployment. We recognize that any such method for controllable generation can be used to generate harmful samples by malicious actors. 
However, \gls{method} enables the research community to better understand properties of generative models and make them safer, which we believe will ultimately outweigh these harms.

\section{Acknowledgments}
The authors would like to acknowledge Stefan Andreas Baumann, Yunfan Zhang, Anshuk Uppal, Mark Goldstein, and Eric Horvitz for their valuable feedback.

This work was partly supported by the NIH/NHLBI Award R01HL148248, NSF Award 1922658 NRT-HDR: FUTURE Foundations, Translation, and Responsibility for Data Science, NSF CAREER Award 2145542, ONR N00014-23-1-2634, and Apple. Additional support was provided by a Fellowship from the Columbia Center of AI Technology. 
This work was also supported by IITP with a grant funded by the MSIT of the Republic of Korea in connection with the Global AI Frontier Lab International Collaborative Research.

\bibliography{fkd_steering}
\bibliographystyle{unsrtnat}

\appendix 

\section{Text to Image Experiments}\label{sec:t2i_appendix}
\begin{table}[h]
    \centering    
    \begin{tabular}{lccccc}
    \toprule
       \textbf{Model}  & \textbf{Params} & \textbf{Base}${(k=1)}$ & \textbf{Base}${(k=4)}$ & \textbf{FK}$(k=4)$ & \textbf{FK}$(k=4, \text{parallel})$ \\
    \midrule       
        SD v1.4/v1.5 & 860M & $2.4$s & $7.3$s & $8.1$s & $5.0$s  \\
        SD v2.1 & 865M & $4.6$s  & $15.6$s & $17.4$s & $9.1$s  \\
        SDXL & 2.6B & $11.5$s   & $42.3$s & $43.5$s & $21.7$s  \\
    \bottomrule        
    \end{tabular}
    \caption{\textbf{Parameter counts and timing}. In this table, we provide inference timing for text-to-image diffusion models with \gls{method}. We include results for \gls{method} on a single NVIDIA-A100 GPU and a two-device parallel implementation.}
    \label{tab:param_count}
\end{table}

In this section, we explore the effect of $\lambda$ and the resampling schedule on particle diversity for text-to-image generation. Similar to \citet{domingo2024adjoint}, we diversity of generations using the CLIP \citep{radford2021learning} encoder $f_\theta$, so given $k$ $\{\mbx_0^i\}_{i=1}^{k}$ particles we measure:
\begin{align}
   \text{CLIP-Div}\left( \{\mbx_0^i\}_{i=1}^{k} \right) := \sum_{i=1}^k \sum_{j=i}^k \frac{2}{k (k-1)} \norm{f_\theta(\mbx_0^i) - f_\theta(\mbx_0^j)}_2^2 .
\end{align}
Similar to \cref{sec:sample_quality}, we use the stable diffusion text-to-image models \citep{rombach2022high} with the ImageReward human preference score \citep{xu2024imagereward} as the reward function. Here we use the difference potential. 

We evaluate \gls{method} with different values of $\lambda$ and different resampling schedules, $[0, 20, 40, 60, 80]$ and $[0, 70, 75, 80, 85, 90]$. In \cref{tab:geneval_diversity}, we observe that for all values of $\lambda$ and the resampling schedule, the GenEval score of \gls{method} outperforms the base model. However, for lower values of $\lambda$, the CLIP diversity score is significantly higher, implying higher particle diversity. Similarly, in \cref{tab:ir_diversity}, we observe that for higher values of $\lambda$, the human preference scores are higher, while the particle diversity is lower. 

\begin{table}[h!]
\centering
\begin{tabular}{llccc}
\toprule
\textbf{Model} & \textbf{Sampler} & \textbf{Schedule} & \textbf{CLIP Div.} & \textbf{GenEval Score} \\
\midrule
SD v1.4 & FK($k=4$, $\lambda = 10$) & 5-30-5 & 0.1437 & 0.4814 \\
SD v1.4 & FK($k=4$, $\lambda = 10$) & 20-80-20 & 0.1050 & 0.5258 \\
SD v1.4 & FK($k=4$, $\lambda = 2$) & 5-30-5 & 0.2321 & 0.4975 \\
SD v1.4 & FK($k=4$, $\lambda = 2$) & 20-80-20 & 0.2239 & 0.4910 \\
\midrule
SD v1.4 & base ($k=4$) & - & 0.3158 & 0.4408 \\
\midrule
SD v1.5 & FK($k=4$, $\lambda = 10$) & 5-30-5 & 0.1459 & 0.4861 \\
SD v1.5 & FK($k=4$, $\lambda = 10$) & 20-80-20 & 0.1038 & 0.5224 \\
SD v1.5 & FK($k=4$, $\lambda = 2$) & 5-30-5 & 0.2330 & 0.4854 \\
SD v1.5 & FK($k=4$, $\lambda = 2$) & 20-80-20 & 0.2252 & 0.5114 \\
\midrule
SD v1.5 & base ($k=4$) & - & 0.3115 & 0.4483 \\
\midrule
SD v2.1 & FK($k=4$, $\lambda = 10$) & 5-30-5 & 0.1259 & 0.5523 \\
SD v2.1 & FK($k=4$, $\lambda = 10$) & 20-80-20 & 0.1061 & 0.5783 \\
SD v2.1 & FK($k=4$, $\lambda = 2$) & 5-30-5 & 0.2051 & 0.5607 \\
SD v2.1 & FK($k=4$, $\lambda = 2$) & 20-80-20 & 0.2213 & 0.5587 \\
\midrule
SD v2.1 & base ($k=4$) & - & 0.2948 & 0.5104 \\
\midrule
SDXL & FK($k=4$, $\lambda = 10$) & 5-30-5 & 0.1182 & 0.6056 \\
SDXL & FK($k=4$, $\lambda = 10$) & 20-80-20 & 0.1055 & 0.6034 \\
SDXL & FK($k=4$, $\lambda = 2$) & 5-30-5 & 0.1816 & 0.5863 \\
SDXL & FK($k=4$, $\lambda = 2$) & 20-80-20 & 0.2111 & 0.5857 \\
\midrule
SDXL & base ($k=4$) & - & 0.2859 & 0.5571 \\
\bottomrule
\end{tabular}
\caption{\textbf{Effect of $\lambda$ and resampling schedule on diversity.} Here we report GenEval scores of all particles generation by \gls{method} to show that prompt fidelity increases for all particles. Moreover, we notice that lower values of $\lambda$ can also be used to generate diverse particles.}\label{tab:geneval_diversity}
\end{table}

\begin{table}[h!]
\centering
\begin{tabular}{llccccc}
\toprule
\textbf{Model} & \textbf{Sampler} & \textbf{Schedule} & \textbf{IR (Mean / Max)} & \textbf{HPS (Mean / Max)} & \textbf{CLIP Div.} \\
\midrule
SD v1.4 & base ($k = 4$) & - & 0.234 (0.800) & 0.245 (0.256) & 0.348 \\
SD v1.4 & FK ($k = 4, \lambda = 10.0$) & 5-30-5 & 0.506 (0.783) & 0.251 (0.255) & 0.193 \\
SD v1.4 & FK ($k = 4, \lambda = 10.0$) & 20-80-20 & 0.811 (0.927) & 0.258 (0.259) & 0.091 \\
SD v1.4 & FK ($k = 4, \lambda = 1.0$) & 20-80-20 & 0.502 (0.763) & 0.252 (0.256) & 0.173 \\
SD v1.4 & FK ($k = 4, \lambda = 1.0$) & 5-30-5 & 0.368 (0.723) & 0.248 (0.254) & 0.236 \\
\midrule
SD v2.1 & base ($k = 4$) & - & 0.372 (0.888) & 0.253 (0.263) & 0.318 \\
SD v2.1 & FK ($k = 4, \lambda = 1.0$) & 5-30-5 & 0.582 (0.835) & 0.258 (0.261) & 0.180 \\
SD v2.1 & FK ($k = 4, \lambda = 10.0$) & 20-80-20 & 0.891 (1.006) & 0.264 (0.266) & 0.087 \\
SD v2.1 & FK ($k = 4, \lambda = 1.0$) & 20-80-20 & 0.579 (0.826) & 0.257 (0.261) & 0.164 \\
\midrule
SDXL & base ($k = 4$) & - & 0.871 (1.236) & 0.289 (0.296) & 0.248 \\
SDXL & FK ($k = 4, \lambda = 10.0$) & 5-30-5 & 1.032 (1.186) & 0.293 (0.295) & 0.123 \\
SDXL & FK ($k = 4, \lambda = 10.0$) & 20-80-20 & 1.211 (1.298) & 0.296 (0.297) & 0.071 \\
\bottomrule
\end{tabular}
\caption{\textbf{Effect of $\lambda$ and resampling schedule on diversity.} Here we report ImageReward and HPS scores of all particles generation by \gls{method} to show that sample quality increases for all particles. Moreover, we notice that lower values of $\lambda$ can also be used to generate diverse particles.}\label{tab:ir_diversity}
\end{table}

\section{Text Experiments}
\label{sec:text_q_appendix}
\label{sec:text_tox_appendix}

For all text experiments, we use publicly available SSD-LM\footnote{\url{https://huggingface.co/xhan77/ssdlm}}, MDLM\footnote{\url{https://huggingface.co/kuleshov-group/mdlm-owt}}, and GPT2-Medium\footnote{\url{https://huggingface.co/openai-community/gpt2-medium}} checkpoints. For both text experiments, we generate sequences of length $50$, conditioned on the prompts used by \citet{han2023ssdlmsemiautoregressivesimplexbaseddiffusion} to evaluate controllable text generation. We generate $20$ continuations for each of the $15$ prompts.

\begin{table*}
\begin{center}
\begin{small}
\begin{tabular}{lccc}
\toprule
\textbf{Model + Sampler} & \textbf{Toxic} $\uparrow$ & \textbf{Toxic (Holdout)} $\uparrow$ & \textbf{PPL (GPT2-XL)} $\downarrow$  \\
\midrule
{SSD-LM} & 0.4\% & 1.2\% & \textbf{23.2}  \\
{SSD-LM} (learned $\nabla$ guidance) & 36.5\% & 40.3\% & 43.3  \\
{SSD-LM} ($\nabla$ guidance) & 22.3\% & 22.6\% & 40.3   \\
MDLM & 0.3\% & 1.9\% & 85.3  \\
\midrule
\multicolumn{4}{l}{{SSD-LM (no gradients)}} \\
\midrule
{{BoN(4)}} & 1.6\% & 4.8\% & 21.9  \\
{{BoN(8)}} & 5.0\% & 8.1\% & 23.0  \\
\textbf{{{FK($k=4$)}}} & 8.4\% & 14.0\%  & 22.5 \\
\textbf{{{FK($k=4$, learned $r_\phi$)}}} & 15.2\% & 19.6\%  & 26.3  \\
\textbf{{{FK($k=8$)}}} & 25.0\% & 29.7\%  & 23.9 \\
\textbf{{{FK($k=8$, learned $r_\phi$)}}} & {39.0\%} & {38.0\%}  & 26.9  \\
\midrule
\multicolumn{4}{l}{{SSD-LM (with gradients)}} \\
\midrule
\textbf{{{FK($k=4$, gradients from learned $r_\phi$)}}} & \textbf{55.6}\% & \textbf{56.3}\%  & 36.0  \\
\bottomrule
\end{tabular}
\caption{\textbf{Toxicity results}. We evaluate the toxicity of the generated samples with (a) the classifier used for steering and (b) a separate holdout classifier, we also report GPT2-XL perplexity. Results are averaged over three seeds.}
\label{tab:reward_gradient_guided_toxicity}
\end{small}
\end{center}
\end{table*}

\subsection{Baselines}

Following \citet{han2023ssdlmsemiautoregressivesimplexbaseddiffusion}, for SSD-LM we iteratively generate these continuations in blocks of $25$. 
Except for our $T=5000$ quality experiment, we default to $T=500$ for all SSD-LM experiments, and follow the multi-hot sampling procedure, with a top-p $ = 0.20$ \citep{han2023ssdlmsemiautoregressivesimplexbaseddiffusion}. For toxicity gradient guidance, we set the learning rate $ = 2000$. For MDLM, we condition on each prompt by prefilling the prompt tokens at inference time. The model is trained to generate tokens in blocks of $1024$. For consistency, we only consider the first $50$ tokens of each generated sample, after re-tokenizing with the SSD-LM tokenizer. We use $1000$ steps for all MDLM experiments. For the GPT2-Medium baseline, we generate all samples with top-p $=0.95$ and temperature $=1.0$.

\subsection{\gls{method} Details}

For all \gls{method} text experiments, we set $\lambda=10.0$ and use the difference of rewards potential. We resample $50$ times for each inference: at every $10$ steps for SSD-LM and every $20$ steps for MDLM.  To convert intermediate SSD-LM states to text, we sample tokens from the logit estimate, $\widehat{\mbx}_t$, with top-p $ = 0.20$.  To convert intermediate MDLM states to text, we sample the masked tokens from the multinomial distribution given by $\widehat{\mbx}_t$. By default, we sample one intermediate text for SSD-LM, and four texts for MDLM. Rewards are averaged over these samples. For \textit{Improved} \gls{method} with MDLM, we sample and evaluate $16$ intermediate texts, rather than $4$. 

For \textit{Improved} \gls{method} with SSD-LM, we take the more involved approach of fine-tuning the off-the-shelf toxicity classifier on intermediate states, $\widehat{x}_t$. To build a training dataset, we used reward toxicity classifier to identify $26$K non-toxic and $26$K toxic texts from the OpenWebText corpus \citep{Gokaslan2019OpenWeb}. We then applied the SSD-LM forward process $q$ to noise the text to random timestep $t$, and then use the base model to infer $\widehat{x}_t$. We then fine-tune the off-the-shelf reward classifier to estimate the toxicity probability of the original text given the intermediate text. 

We fine-tune three reward models for different SSD-LM time-step ranges:
\[ t \in [500,300), [300,200), [200,100) \]
We train with batch size $ = 16$ and learning rate $ = 5e-7$, using a constant learning rate with $50$ warm-up steps. We train with cross entropy loss, and use a weighting ($0.99$ non-toxic, $0.01$ toxic), due to the rarity of toxicity in the original data distribution. For the gradient-based guidance baseline for SSD-LM,  we use the default guidance scale from \citet{han2023ssdlmsemiautoregressivesimplexbaseddiffusion}\footnote{Provided in private communication by the authors of \citet{han2023ssdlmsemiautoregressivesimplexbaseddiffusion}.}.

\section{Feynman Kac IPS discussion}\label{sec:discrete_proofs}
\subsection{Choice of proposal distribution}\label{sec:twisted_transition}
Here we discuss various choices for twisting the transition kernel towards high reward samples:
\begin{itemize}[leftmargin=*]
    \item \textbf{Gradient-based guidance}: For continuous-state models and differentiable rewards, we can use gradient's from the reward \citep{sohl2015deep,song2020score,bansal2023universal,wu2024practical} to guide the sampling process. Suppose $p_\theta(\mbx_t \mid \mbx_{t+1}, \mbc) = \cN(\mu_\theta(\mbx_t, \mbc), \sigma^2_\theta I_d)$, then we can \textit{twist} the transition kernel using reward gradients:
    \begin{align}
        \cN\left( \mu_\theta(\mbx_t, \mbc) + \sigma^2_\theta \lambda \nabla_{\mbx_t} r_\phi(\mbx_t, \mbc), \sigma^2_\theta \right) ,
    \end{align}
    where $r_\phi$ is the intermediate reward, either learned or evaluated at the reward on the denoised state $r(\mbx_0 = \widehat{\mbx}_t)$.
    \item \textbf{Discrete normalization}: For discrete diffusion models, such as \gls{mdlm} \citep{sahoo2024simple,shi2024simplified}, we can also estimate the normalization constant:
    \begin{align}
         \sum_{\mbx_t} p_\theta(\mbx_t \mid \mbx_{t+1}, \mbc) G_t(\mbx_T, \dots, \mbx_t, \mbc)
    \end{align}
    and sample from $\pfk(\mbx_t \mid \mbx_{t+1}, \dots, \mbx_T) \propto p_\theta(\mbx_t \mid \mbx_{t+1}) G_t(\mbx_T, \dots, \mbx_t)$. 
    \end{itemize}
However, such methods for twisting the transition kernel can lead to increased sampling time compared to sampling from the \textit{base} model $p_\theta$.

\subsection{How existing work fits into \gls{method}}\label{sec:existing_work}
\gls{tds} \citep{wu2024practical} uses \gls{smc} to do conditional sampling with a marginally trained model and a differentiable reward. They make the choices:
\begin{itemize}
    \item \textbf{Potential.} $G_t(\mbx_t, \mbx_{t+1}) = \exp(\lambda(r(\mbx_t) - r(\mbx_{t+1})))$, where the reward is computed on the denoised state $r(\mbx_t) = r(\mbx_0 = \widehat{\mbx}_t)$.
    \item \textbf{Proposal generator.} They use classifier-guidance to approximate the conditional score model $s_\theta(\mbx_t, t, y) \approx s_\theta(\mbx_t, t) + \nabla_{\mbx_t} \log p_\theta(y \mid \mbx_0 = \widehat{\mbx}_t(\mbx_t, t))$ and use the following proposal generator $\tau(\mbx_t \mid \mbx_{t+1})$:
    \begin{align}
        \tau(\mbx_t \mid \mbx_{t+1}) &= \textsc{N}(\Delta t [f - gg^\top s_\theta(\mbx_t, t, y)], g(t) \Delta t )
    \end{align}        
\end{itemize}
\gls{method} allows for a more flexible use of potentials $G_t$, as well as proposal generators. For instance, \citet{nichol2021glide} show that conditionally trained scores outperform classifier-guidance even when the classifier is trained on noisy states $\mbx_t$. However, as shown by \citet{ghosh2024geneval}, conditionally trained models still have failure modes. Therefore, we demonstrate how particle based methods can be used to improve the performance of conditionally trained models as well. Furthermore, \gls{method} allows these methods to be applied to discrete-space diffusions as well as non-differentiable rewards. 

\gls{svdd} is a particle-based method which instead of using \gls{smc}, uses  a nested importance sampling algorithm (see algorithm 5 of \citet{naesseth2019elements}). \gls{svdd} makes the following choices:
\begin{itemize}
    \item \textbf{Potential.} Similar to \gls{tds}, they use the potential $G_t = \exp(\lambda(r(\mbx_t) - r(\mbx_{t+1})))$ where $r(\mbx_t)$ can be off-the-shelf like \gls{tds} or learned from model samples.  
    \item \textbf{Sampler.} \gls{svdd} uses the base model as the proposal generator and generates $k$ samples at each step, selects a \textit{single sample} using importance sampling and makes $k$ copies of it for the next step.     
\end{itemize}
With $\lambda = \infty$, \gls{svdd} is equivalent to doing best-of-$n$ at each step, since the authors recommend sampling from $p_\text{target}(\mbx_0) \propto \lim_{\lambda \rightarrow \infty} p_\theta(\mbx_0) \exp(\lambda r(\mbx_0))$. We note that similar to \gls{svdd}, $\pfk{0}$ can be sampled using nested importance sampling.

\subsection{Adaptive Resampling}\label{sec:adaptive_resampling}
Following \citet{naesseth2019elements,wu2024practical}, we use adaptive resampling to increase diversity of samples. Given $k$ particles $\mbx_t^i$ and their potentials $G_t^i$, we define the effective sample size (\textsc{ess}):
\begin{align}
    \text{\textsc{ess}}_t = \frac{1}{\sum_{i=1}^k \left(\widehat{G}_t^{i} \right)^2}
\end{align}
where $\widehat{G}$ refers to the normalized potentials and $\text{\textsc{ess}}_t \in [1, k]$. If $\text{\textsc{ess}}_t < \frac{k}{2}$, then we skip the resampling step. This encourages particle diversity.  

\section{\gls{method} samples}\label{sec:samples}
In this section, we show the effect of various sampling parameters, such as potentials, the temperature parameter $\lambda$, number of sampling steps, etc. on the diversity of samples. We use the stable diffusion XL-base (SDXL) as the base model and proposal generator and the ImageReward \citep{xu2024imagereward} human preference score model as the reward function. We also use adaptive resampling introduced in \cref{sec:adaptive_resampling}. We compare \gls{method} against generating $k$ independent samples, using the same seed for generation, thus providing a counterfactual generation.

\begin{figure}[t]
    \centering
\includegraphics[width=0.8\linewidth]{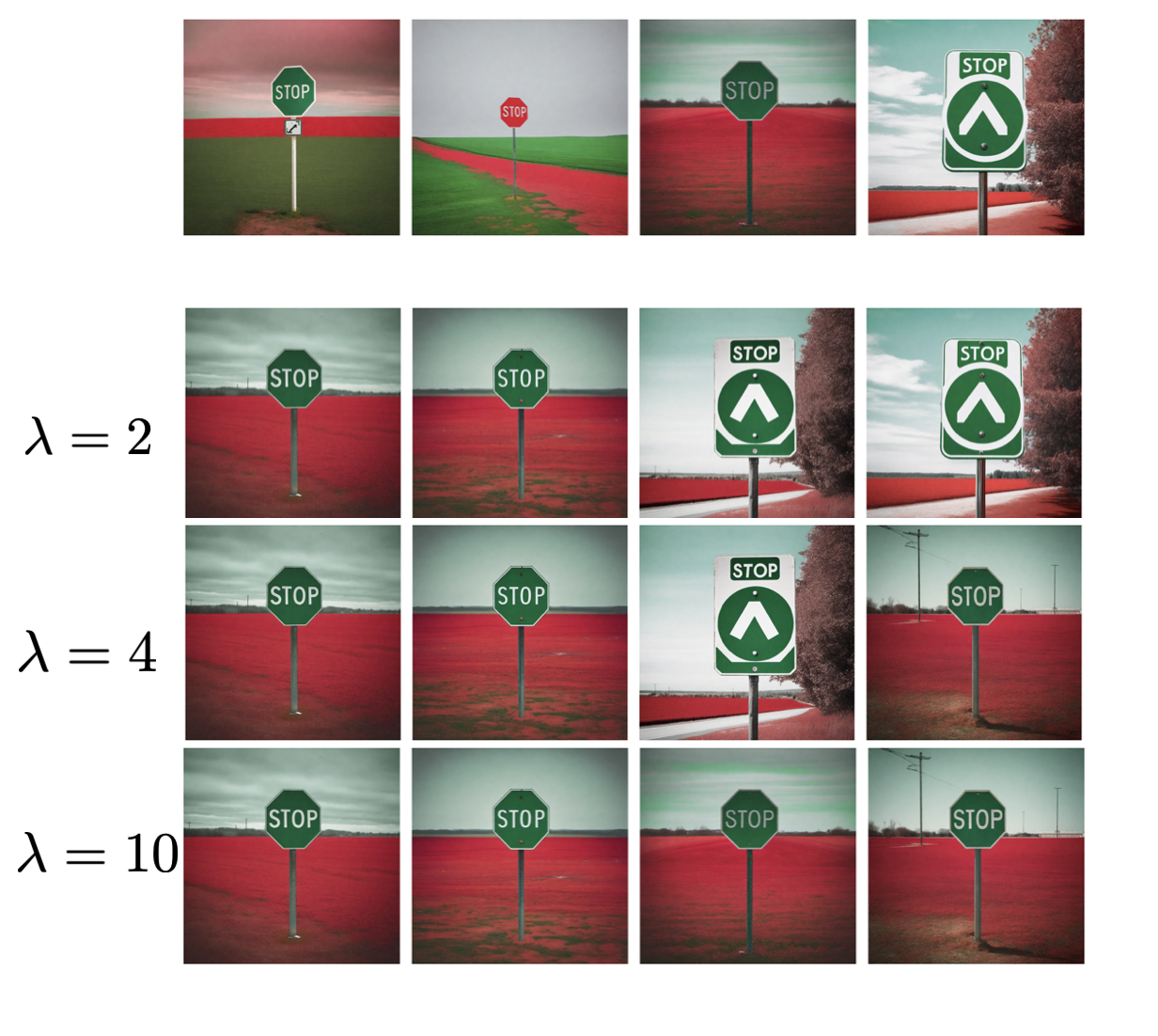}
    \caption{\textbf{Effect on $\lambda$ on diversity:} In the top panel, we plot $4$ independent samples from the base model and in the bottom 3 panels, we have the \gls{method} particles for varying values of $\lambda$. We observe that increasing $\lambda$ leads to a decrease in diversity, at the cost of higher prompt fidelity and improved aesthetic quality, compared to the first row which has $4$ independent samples. Caption: \textit{a green stop sign in a red field}}
    \label{fig:effect_of_lambda}
\end{figure}

\begin{figure}[t]
    \centering    \includegraphics[width=1.0\linewidth]{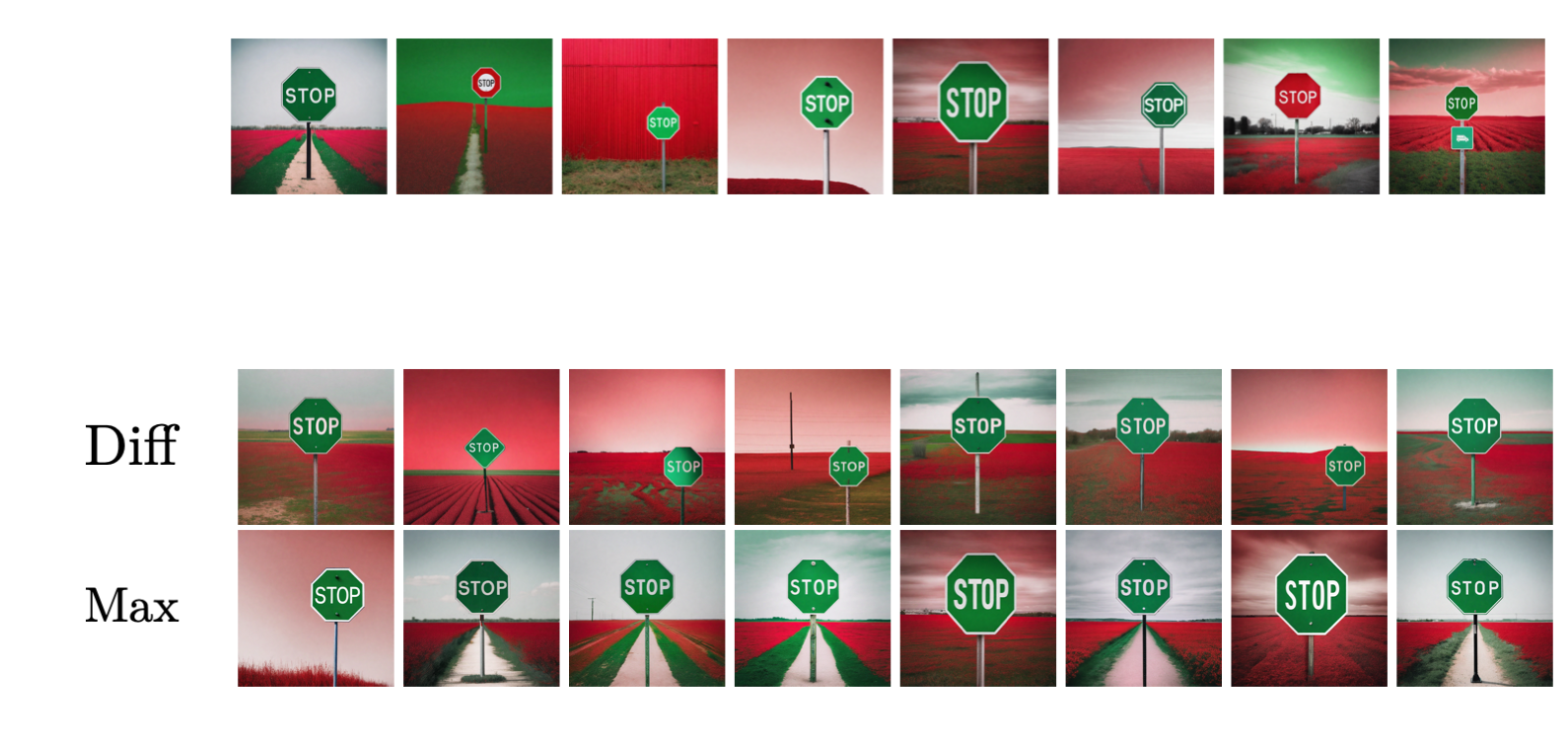}
    \caption{\textbf{Max versus Difference potential}: In the top row, we plot $8$ independent samples from the base model and in the bottom two rows, we have the \gls{method} particles for the max and difference potentials. Using the max potential reduces diversity compared to the difference potential. However, we note that by increasing the number of sampling steps, the diversity of the samples can be increased. Caption: \textit{a green stop sign in a red field}}
    \label{fig:effect_of_potential}
\end{figure}

\begin{figure}[t]
    \centering
    \includegraphics[width=0.5\linewidth]{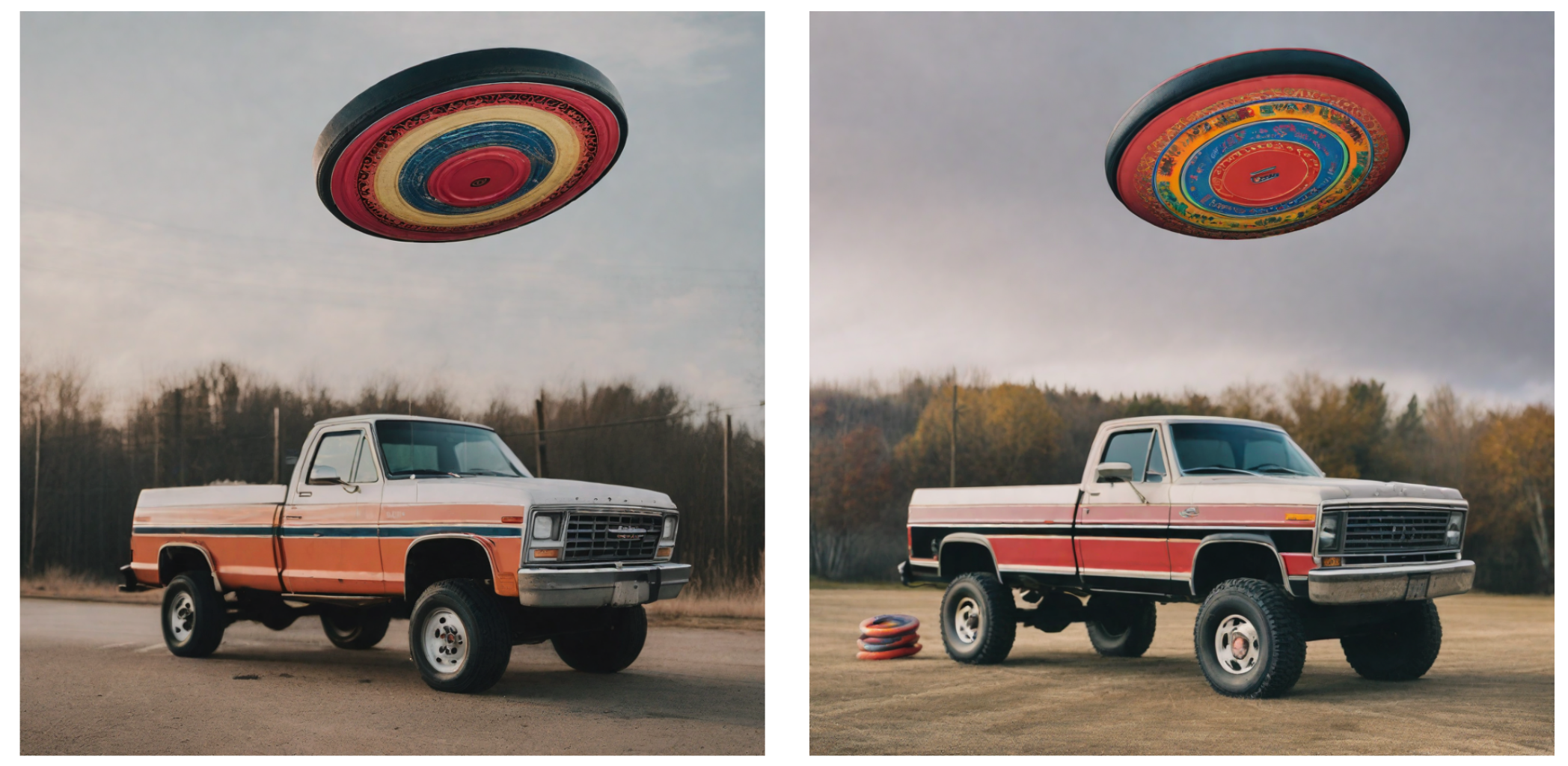}
    \caption{\textbf{Effect of interval resampling:} While the overall diversity is reduced, using interval resampling encourages diversity. Caption: \textit{a photo of a frisbee above a truck}}
    \label{fig:effect_of_interval_resampling}
\end{figure}

\begin{figure}
    \centering
    \includegraphics[width=0.9\linewidth]{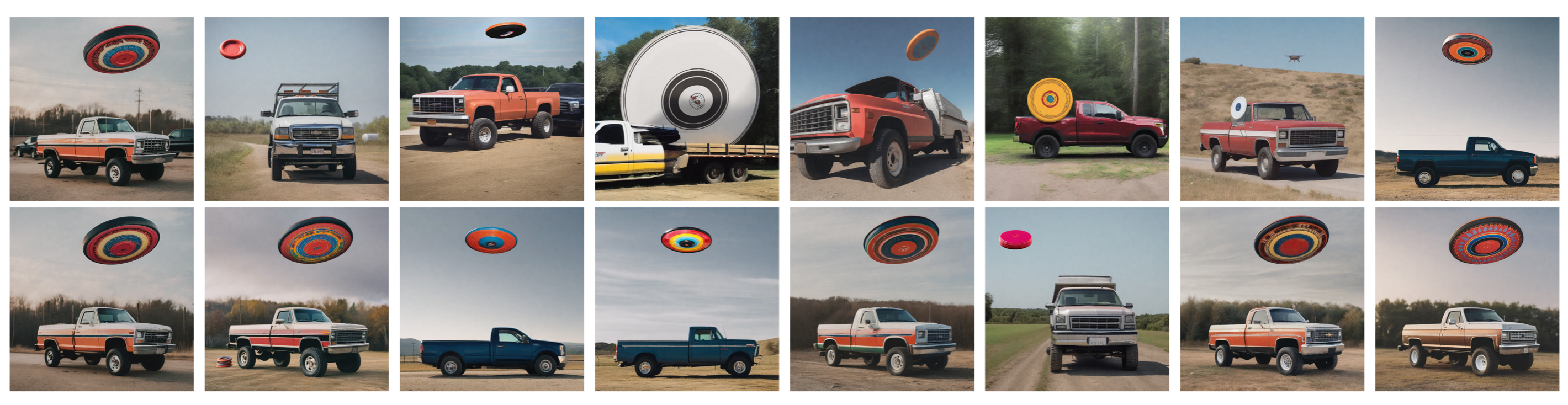}
    \caption{\textbf{Increased prompt fidelity}: In this generation, we compare $k=8$ independent samples (top panel) versus $k=8$ samples from \gls{method} (bottom panel). \gls{method} selects samples which follow the prompt. Caption: \textit{a photo of a frisbee above a truck}}
    \label{fig:spatial_intelligence}
\end{figure}

\begin{itemize}
    \item \textbf{Effect of $\lambda$}: The parameter $\lambda$ is used to define the target distribution:
    \begin{align}
        \ptarget(\mbx_0) &= \frac{1}{\mbZ} p_\theta(\mbx_0) \exp(\lambda r(\mbx_0)),
    \end{align}
    therefore, higher values of $\lambda$ upweight higher reward samples $\mbx_0$. Similarly, the potentials also use $\lambda$ which affects resampling. We generate $k=4$ samples from the SDXL using \gls{method} as well as $k=4$ independent samples using the max potential. In \cref{fig:effect_of_lambda}, we observe that using \gls{method} improves prompt fidelity, and higher values of $\lambda$ improve fidelity at the cost of particle diversity. 

    \item \textbf{Effect of potential}: In \cref{fig:effect_of_potential}, we observe that \gls{method} with the max potential reduces diversity compared to the difference potential. Here we use $\lambda = 2$ and generate $k=8$ samples using the max and difference potential. 
    \item \textbf{Effect of sampling steps.}  In \cref{fig:effect_of_potential}, we observe that diversity can be increased by increasing the number of sampling steps from 100 to 200. Here we use $[180, 160, 140, 120, 0]$ and  $[80, 60, 40, 20, 0]$ as the resampling interval. We note that even if the samples $\mbx_0$ share the same particle as parent, there is diversity in the final samples. 
    \item \textbf{Effect of interval resampling}: In \cref{fig:effect_of_interval_resampling}, we show that using interval resampling even with $100$ sampling steps produces diversity in samples. For comparison, see \cref{fig:spatial_intelligence} for the independent versus \gls{method} generations.
\end{itemize}

\end{document}